\documentclass[lettersize,journal]{IEEEtran}
\usepackage{amsmath,amsfonts}
\usepackage{array}
\usepackage[caption=false,font=normalsize,labelfont=sf,textfont=sf]{subfig}
\usepackage{textcomp}
\usepackage{stfloats}
\usepackage{verbatim}
\usepackage{cite}
\usepackage{enumitem}
\usepackage{caption}
\usepackage{times}
\usepackage{epsfig}
\usepackage{graphicx}
\usepackage{amsmath}
\usepackage{amssymb}
\usepackage{eso-pic}
\usepackage{helvet}
\usepackage{courier}
\usepackage{url}
\usepackage{mathrsfs}
\usepackage{multirow}
\usepackage{comment}
\usepackage{marvosym}
\usepackage{bbm}
\usepackage{enumerate}
\usepackage{soul}
\usepackage[utf8]{inputenc}
\usepackage{amsfonts}
\usepackage{nicefrac}
\usepackage{microtype}
\usepackage{booktabs} 
\usepackage{algorithm}
\usepackage{algorithmicx}
\usepackage{algpseudocode}
\usepackage{makecell}
\usepackage{multirow}

\usepackage{bbding}
\usepackage[numbers]{natbib}
\usepackage[breaklinks=true,bookmarks=false,colorlinks]{hyperref}
\usepackage{cleveref}
\usepackage[normalem]{ulem}

\usepackage{color, colortbl}
\usepackage{xcolor}

\definecolor{Gray}{gray}{0.9}
\newcommand\crule[3][black]{\textcolor{#1}{\rule{#2}{#3}}}
\definecolor{roadcolor}{RGB}{234,51,246}
\definecolor{sidewalkcolor}{RGB}{68,8,72}
\definecolor{parkingcolor}{RGB}{241,156,249}
\definecolor{othergroundcolor}{RGB}{160,32,76}
\definecolor{buildingcolor}{RGB}{246,202,69}
\definecolor{carcolor}{RGB}{111,149,238}
\definecolor{truckcolor}{RGB}{74,32,172}
\definecolor{bicyclecolor}{RGB}{136,227,242}
\definecolor{motorcyclecolor}{RGB}{37,59,146}
\definecolor{othervehiclecolor}{RGB}{96,81,242}
\definecolor{vegetationcolor}{RGB}{79, 173, 50}
\definecolor{trunkcolor}{RGB}{126, 65, 22}
\definecolor{terraincolor}{RGB}{171, 238, 105}
\definecolor{personcolor}{RGB}{234, 60, 49}
\definecolor{bicyclistcolor}{RGB}{234, 66, 195}
\definecolor{motorcyclistcolor}{RGB}{138, 42, 90}
\definecolor{fencecolor}{RGB}{238, 128, 69}
\definecolor{polecolor}{RGB}{252, 241, 161}
\definecolor{trafficsigncolor}{RGB}{233, 51, 35}
\definecolor{color1}{RGB}{176, 36, 24}
\definecolor{color2}{RGB}{119,185,0}
\definecolor{color3}{RGB}{0, 0, 200}
\definecolor{colorofteaser}{RGB}{176, 36, 24}
\definecolor{LightGrey}{rgb}{.9,.9,.9}
\definecolor{White}{rgb}{1.,0.,1.}
\definecolor{first}{rgb}{.8,.0,.0}
\definecolor{second}{rgb}{.0,.6,.0}
\definecolor{third}{rgb}{.0,.0,.8}
\definecolor{ceiling}{RGB}{214,  38, 40}   %
\definecolor{floor}{RGB}{43, 160, 4}     %
\definecolor{wall}{RGB}{158, 216, 229}  %
\definecolor{window}{RGB}{114, 158, 206}  %
\definecolor{chair}{RGB}{204, 204, 91}   %
\definecolor{bed}{RGB}{255, 186, 119}  %
\definecolor{sofa}{RGB}{147, 102, 188}  %
\definecolor{table}{RGB}{30, 119, 181}   %
\definecolor{tvs}{RGB}{160, 188, 33}   %
\definecolor{furniture}{RGB}{255, 127, 12}  %
\definecolor{objects}{RGB}{196, 175, 214} %
\definecolor{car}{rgb}{0.39215686, 0.58823529, 0.96078431}
\definecolor{bicycle}{rgb}{0.39215686, 0.90196078, 0.96078431}
\definecolor{motorcycle}{rgb}{0.11764706, 0.23529412, 0.58823529}
\definecolor{truck}{rgb}{0.31372549, 0.11764706, 0.70588235}
\definecolor{other-vehicle}{rgb}{0.39215686, 0.31372549, 0.98039216}
\definecolor{person}{rgb}{1.        , 0.11764706, 0.11764706}
\definecolor{bicyclist}{rgb}{1.        , 0.15686275, 0.78431373}
\definecolor{motorcyclist}{rgb}{0.58823529, 0.11764706, 0.35294118}
\definecolor{road}{rgb}{1.        , 0.        , 1.        }
\definecolor{parking}{rgb}{1.        , 0.58823529, 1.        }
\definecolor{sidewalk}{rgb}{0.29411765, 0.        , 0.29411765}
\definecolor{other-ground}{rgb}{0.68627451, 0.        , 0.29411765}
\definecolor{building}{rgb}{1.        , 0.78431373, 0.        }
\definecolor{fence}{rgb}{1.        , 0.47058824, 0.19607843}
\definecolor{vegetation}{rgb}{0.        , 0.68627451, 0.        }
\definecolor{trunk}{rgb}{0.52941176, 0.23529412, 0.        }
\definecolor{terrain}{rgb}{0.58823529, 0.94117647, 0.31372549}
\definecolor{pole}{rgb}{1.        , 0.94117647, 0.58823529}
\definecolor{traffic-sign}{rgb}{1.        , 0.        , 0.    }

\definecolor{barrier1}{RGB}{112,128,144}
\definecolor{bicycle1}{RGB}{220,20,60}
\definecolor{bus1}{RGB}{255, 127, 80}
\definecolor{car1}{RGB}{255, 158, 0}
\definecolor{const. veh.1}{RGB}{233, 150, 70}
\definecolor{motorcycle1}{RGB}{255,61,99}
\definecolor{pedestrian1}{RGB}{0,0,230}
\definecolor{traffic cone1}{RGB}{47,79,79}
\definecolor{trailer1}{RGB}{255,140,0}
\definecolor{truck1}{RGB}{255,99,71}
\definecolor{drive. suf.1}{RGB}{0,207,191}
\definecolor{other flat1}{RGB}{175,0,75}
\definecolor{sidewalk1}{RGB}{75,0,75}
\definecolor{terrain1}{RGB}{112,180,60}
\definecolor{manmade1}{RGB}{222,184,135}
\definecolor{vegetation1}{RGB}{0,175,0}

\definecolor{nbarrier}{RGB}{255, 120, 50}
\definecolor{nbicycle}{RGB}{255, 192, 203}
\definecolor{nbus}{RGB}{255, 255, 0}
\definecolor{ncar}{RGB}{0, 150, 245}
\definecolor{nconstruct}{RGB}{0, 255, 255}
\definecolor{nmotor}{RGB}{200, 180, 0}
\definecolor{npedestrian}{RGB}{255, 0, 0}
\definecolor{ntraffic}{RGB}{255, 240, 150}
\definecolor{ntrailer}{RGB}{135, 60, 0}
\definecolor{ntruck}{RGB}{160, 32, 240}
\definecolor{ndriveable}{RGB}{255, 0, 255}
\definecolor{nother}{RGB}{139, 137, 137}
\definecolor{nsidewalk}{RGB}{75, 0, 75}
\definecolor{nterrain}{RGB}{150, 240, 80}
\definecolor{nmanmade}{RGB}{213, 213, 213}
\definecolor{nvegetation}{RGB}{0, 175, 0}

\usepackage{soul}

\begin{document}

\title{ 
Hierarchical Context Alignment with \\Disentangled Geometric and Temporal \\Modeling for Semantic Occupancy Prediction}

\author{Bohan Li,~Jiajun Deng~\IEEEmembership{Member, IEEE},~Yasheng Sun,~Xiaofeng Wang,\\~Xin Jin,\textsuperscript{\Letter}~\IEEEmembership{Member, IEEE}, ~Wenjun Zeng,~\IEEEmembership{Fellow, IEEE}
\thanks{

Bohan Li is with Shanghai Jiao Tong University, and the Eastern Institute of Technology, Ningbo, China (e-mail: bohan\_li@sjtu.edu.cn).
Xin Jin (corresponding author) is an assistant professor, and Wenjun Zeng is a chair professor at the Ningbo Institute of Digital Twin, Eastern Institute of Technology, Ningbo, China, (e-mail: jinxin@eitech.edu.cn, wenjunzengvp@eitech.edu.cn).

Jiajun Deng is with the University of Adelaide (UoA), Australia (e-mail: jiajun.deng@adelaide.edu.au).	

Yasheng Sun is with the Tokyo Institute of Technology, Tokyo, Japan (e-mail: sun.y.aj@m.titech.ac.jp).

Xiaofeng Wang is with the Institute of Automation, Chinese Academy of Sciences, Beijing, China (e-mail: wangxiaofeng2020@ia.ac.cn).

}
}

\markboth{SUBMITTED TO IEEE Transactions on Pattern Analysis and Machine Intelligence}%
{Shell \MakeLowercase{\textit{et al.}}: A Sample Article Using IEEEtran.cls for IEEE Journals}


\maketitle

\begin{abstract}  
Camera-based 3D Semantic Occupancy Prediction (SOP) is crucial for understanding complex 3D scenes from limited 2D image observations. Existing SOP methods typically aggregate contextual features to assist the occupancy representation learning, alleviating issues like occlusion or ambiguity. However, these solutions often face misalignment issues wherein the corresponding features at the same position across different frames may have different semantic meanings during the aggregation process, which leads to unreliable contextual fusion results and an unstable representation learning process. To address this problem, we introduce a new \textit{Hi}erarchical context alignment paradigm for a more accurate \textit{SOP} \textit{(Hi-SOP)}. Hi-SOP first disentangles the geometric and temporal context for separate alignment, which two branches are then composed to enhance the reliability of SOP. This parsing of the visual input into a local-global alignment hierarchy includes: (I) disentangled geometric and temporal separate alignment, within each leverages depth confidence and camera pose as prior for relevant feature matching respectively; (II) global alignment and composition of the transformed geometric and temporal volumes based on semantics consistency. Our method outperforms SOTAs for semantic scene completion on the SemanticKITTI \& NuScenes-Occupancy datasets and LiDAR semantic segmentation on the NuScenes dataset.
\end{abstract}

 
\section{Introduction}
\IEEEPARstart{C}{omprehending} holistic 3D scenes is crucial for autonomous driving systems, as it significantly influences the planning and obstacle avoidance capabilities for autonomous vehicle safety and efficiency~\cite{li2023front,li2023voxformer,huang2023tri,wei2023surroundocc,li2023delving,luo2024exploring}. However, the limitations of real-world sensors, including restricted fields of view and measurement noise, present substantial challenges. To overcome these difficulties, 3D semantic occupancy prediction (SOP) has been developed to simultaneously infer the geometry and semantics of the scenario from partial observations~\cite{rist2021semantic,roldao2020lmscnet,cao2022monoscene,li2023voxformer,huang2023tri}.\looseness=-1

Given the inherent 3D nature, numerous semantic occupancy prediction (SOP) solutions rely on LiDAR for accurate location measurements~\cite{rist2021semantic,garbade2019two,roldao2020lmscnet}. Although LiDAR provides precise depth information, it inevitably introduces significant cost and manual effort with dense annotations and sophisticated devices. Consequently, it is urgent to explore an efficient approach for precise SOP with a cost-effective scheme. This motivation has prompted the exploration of camera-based solutions, which are characterized by superior deployment efficiency and offer richer visual context, making them a promising alternative for SOP~\cite{cao2022monoscene,wei2023surroundocc,zhang2023occformer,huang2023tri}.\looseness=-1

\begin{figure*}[!h]
\begin{center} 
\vspace{-20pt}
\includegraphics[width=0.7\linewidth]{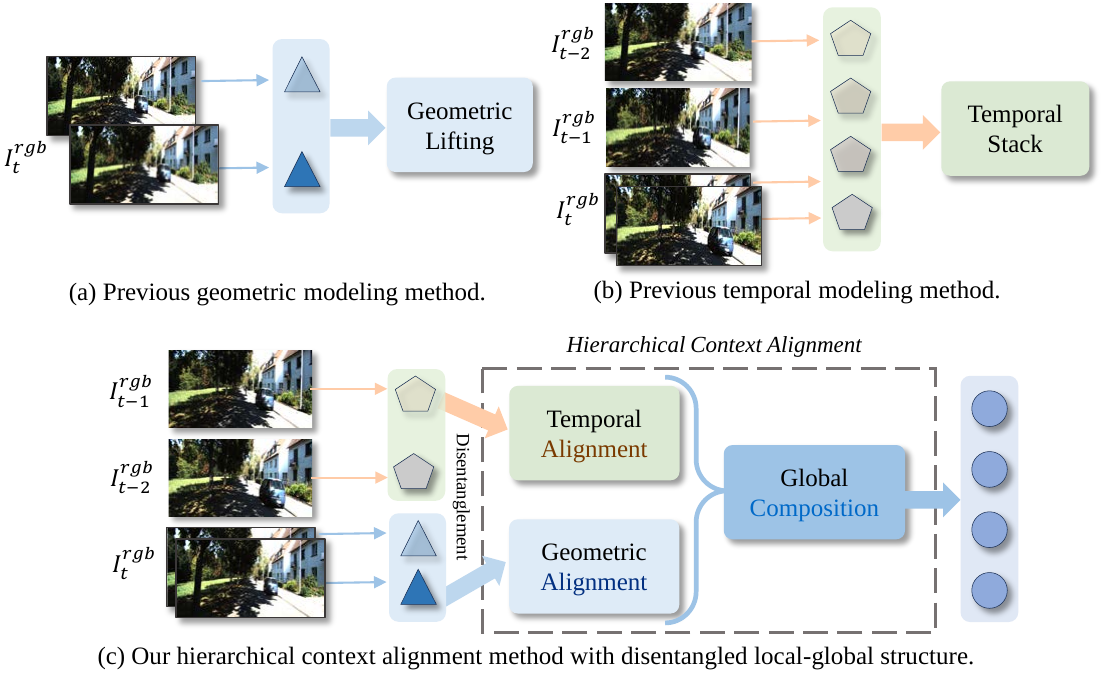}
 \vspace{-5pt}
\caption{
Comparison of our hierarchical context alignment method with prior geometric modeling (\textit{e.g.}, OccFormer~\cite{zhang2023occformer}) and temporal modeling approaches (\textit{e.g.}, VoxFormer-T~\cite{li2023voxformer}) for semantic occupancy prediction. Previous methods handle either geometric lifting or temporal stacking separately, often fusing contexts trivially in a black-box manner, which leads to misalignment and unreliable fusion. In contrast, our approach integrates both geometric and temporal representations in a hierarchically aligned manner, enabling robust contextual composition.}
\label{teaser1}
 \vspace{-15pt}
\end{center}
\end{figure*}

To construct accurate occupancy representations, previous camera-based SOP methods have explored contextual feature aggregation from both geometric and temporal perspectives~\cite{li2024bridging, xue2024bi, song2017semantic, cao2022monoscene, roldao20223d}. As shown in Figure~\ref{teaser1} (a), prior geometric modeling approach (\textit{eg.}, OccFormer~\cite{zhang2023occformer}) typically employs geometric lifting in the voxel feature construction process for image-to-3D transformation. While the temporal modeling approach (\textit{eg.}, VoxFormer-T~\cite{li2023voxformer}) utilizes temporal coherence by stacking multiple historical frames as supplements to the current stereo frame, as illustrated in Figure~\ref{teaser1} (b). Despite these significant contributions, these methods failed to simultaneously address both geometric and temporal aspects and also trivially fused contextual information in a black-box manner~\cite{xue2024bi,li2024bridging}.\looseness=-1

\begin{figure}[!h]
\begin{center} 
\vspace{-0pt}
\includegraphics[width=0.99\linewidth]{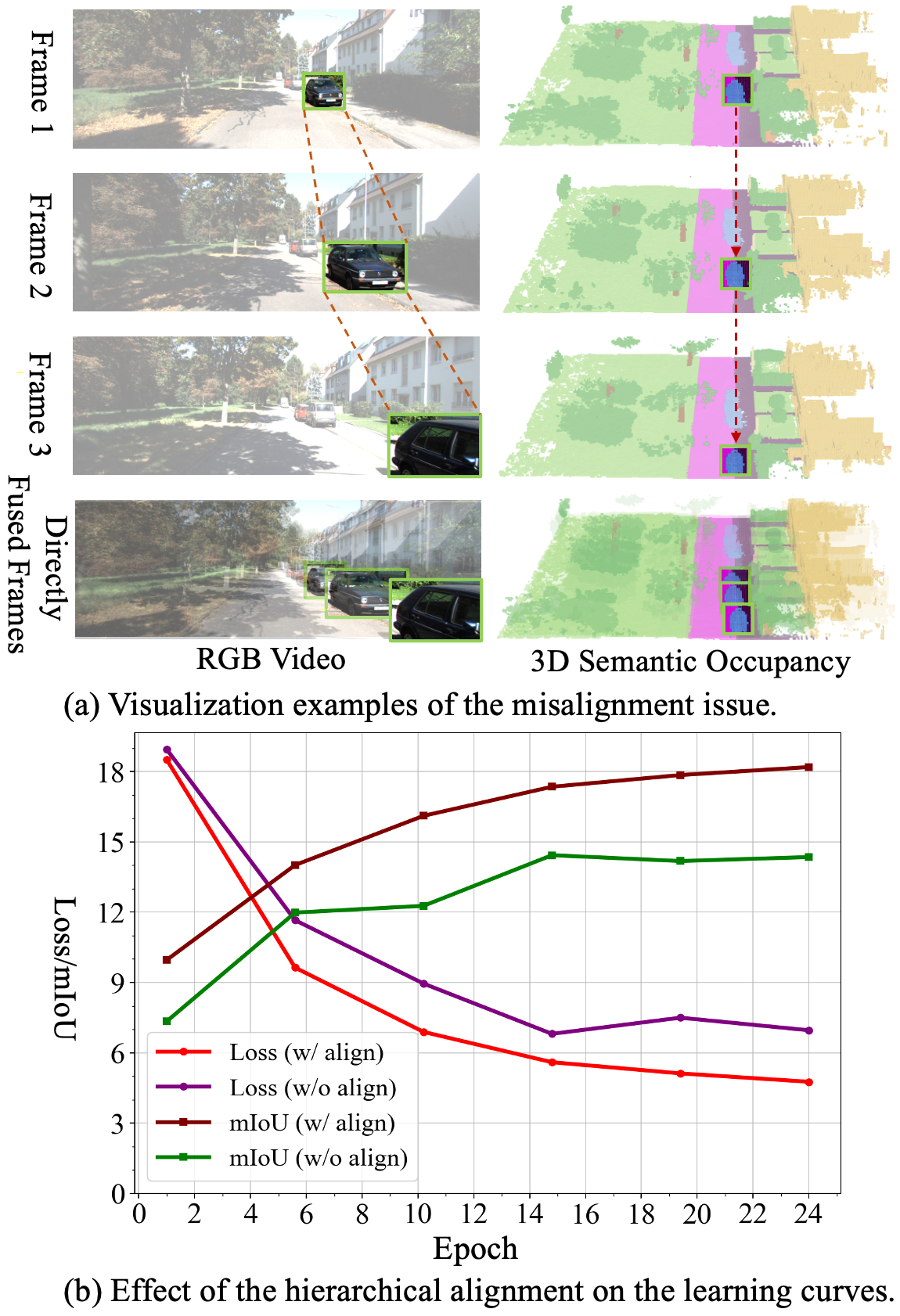}
\vspace{-10pt}
\caption{
(a) Visualization examples of the misalignment issue. The same car appears at different locations across multiple frames. 
Directly fusing contextual information from these frames neglects positional shifts and scale variations, which could lead to ambiguous contextual aggregation and unstable representation learning for semantic occupancy prediction.
(b) The effect of the hierarchical context alignment on the SemanticKITTI validation set. 
We remove both the temporal alignment and the geometric alignment to implement the setting of `w/o align'.
The proposed hierarchical context alignment strategy captures more reliable and comprehensive semantic scenes, and leads to more stable representation modeling in the learning process.\looseness=-1
}
\label{teaser2}
 \vspace{-15pt}
\end{center}
\end{figure}

{As a result, existing SOP solutions inevitably face the misalignment issue during the scene modeling process in the geometric or temporal aspects. Specifically, features corresponding to the same spatial location across different frames may carry inconsistent semantic meanings during aggregation, which can result in fuzzy contextual fusion and unstable representation learning in camera-based visual perception~\cite{song2025traf,pan2025overcoming,yang2024align,park2024adversarial,guo2024damsdet}. 
For instance, as illustrated in Figure~\ref{teaser2}, a moving car may appear at different locations in multiple RGB frames. Directly fusing feature representations extracted from these frames assumes pixel-level correspondence between features from different viewpoints, which is often invalid. This mismatch causes blurred predictive information, as shared semantic content undergoes unpredictable positional shifts that are not consistently aligned across perspectives.
Notably, previous SOP methods~\cite{zhang2023occformer,li2024hierarchical2,li2023voxformer} fail to jointly address geometric and temporal aspects while naively fusing contextual information in a black-box manner, which exacerbates misalignment issues, resulting in unreliable contextual fusion and unstable representation learning.
For instance, as shown in Figure~\ref{teaser1} (a), the geometric modeling presented in OccFormer~\cite{zhang2023occformer} neglects the uncertainty inherent in monocular depth estimation during the voxel feature lifting process. This oversight causes geometric ambiguity when depth information is integrated with the corresponding contextual features. Similarly, Figure~\ref{teaser1}(b) highlights that VoxFormer-T~\cite{li2023voxformer} assumes direct pixel-level correspondence between temporal features from different viewpoints through simple aggregation. However, this assumption disregards positional shifts of shared semantic content across perspectives, yielding blurred predictions and compromising the stability of SOP.}\looseness=-1

{To address these limitations, we propose Hi-SOP, a novel hierarchical context alignment framework that addresses the critical misalignment issue in camera-based semantic occupancy prediction by disentangling and separately aligning geometric and temporal contexts before composition. These aligned contexts are then composed within a unified space to produce reliable semantic occupancy grids.
As depicted in Figure~\ref{teaser1} (c), our method takes advantage of the complementary merits of the geometric and temporal representations (one for the spatial and the other for historic feature perception), and hierarchically aligns them for a reliable context composition.
This hierarchical context alignment with a disentangled local-global structure is composed of two sequential steps: 
(I) individual geometric alignment across frames at different views with optional monocular or stereo depth estimation, and temporal alignment through homography warping and affinity-aware dynamic refinement; 
(II) globally align and compose the geometric and temporal context within a unified space through semantically consistent
transformation and aggregation for a final reliable occupancy prediction.
As shown in Figure~\ref{teaser2}, our proposed hierarchical context alignment strategy Hi-SOP shows promising performance in capturing more reliable and comprehensive semantic scenes, leading to more accurate {prediction} results and more stable learning processes.}\looseness=-1

More specifically, to facilitate reliable geometric alignment, we design a Geometric Confidence-aware Lifting (GCL) module in the geometric alignment branch, which models geometric information with depth distribution confidence awareness before integrating it with corresponding contextual features for voxel feature lifting.
For temporal-wise alignment, we first employ an epipolar homography-warping to explicitly align temporal invariant features and create temporal feature volumes to preserve detailed context. 
To separate critical relevant context from redundant information, we construct a Cross-frame Pattern Affinity (CPA) to measure the contextual relevance, and accordingly based on it refine the dynamic temporal content to compensate for incomplete observations.
Finally, in the composition stage, we propose to globally align the geometric context with the temporal context within a unified space by a Depth-Hypothesis-Based Transformation (DHBT) for semantic-consistent aggregation, which takes the depth hypothesis of the temporal feature volume as the distance axis and {employs} a voxel-pooling operation to splat the volumetric features into the unified space.
{This paper is structured as follows: Section II reviews related work on semantic occupancy prediction, BEV representations, and temporal modeling. Section III details our proposed Hi-SOP framework, including the hierarchical context alignment paradigm, disentangled geometric confidence-aware lifting (GCL) and temporal alignment modules, cross-frame pattern affinity (CPA) measurement, affinity-based dynamic refinement (ADR), and depth-hypothesis-based transformation (DHBT) for global composition. Section IV presents extensive experiments on SemanticKITTI, NuScenes, and NuScenes-Occupancy datasets, validating our method's superiority in semantic scene completion and LiDAR semantic segmentation against state-of-the-art approaches. Ablation studies further analyze key components. Section V concludes the work and discusses future directions.}\looseness=-1

We conduct extensive experiments to validate the advantages of our proposed hierarchical context alignment paradigm for Semantic Occupancy Prediction (SOP) in the semantic scene completion (SSC) and LiDAR semantic segmentation tasks. 
For SSC, our camera-based Hi-SOP outperforms state-of-the-art VoxFormer-T~\cite{li2023voxformer} on the SemanticKITTI~\cite{behley2019semantickitti} benchmark and even surpasses LiDAR-based methods on the NuScenes-Occupancy~\cite{wang2023openoccupancy} benchmark.
We also evaluate our method on the NuScenes~\cite{caesar2020nuscenes} dataset for LiDAR semantic segmentation. Hi-SOP surpasses TPVFormer~\cite{huang2023tri} with a relative improvement of 24.28\% in terms of mIoU. 
This work is an extension version that upgrades our previous ECCV-24 conference paper HTCL~\cite{li2024hierarchical2} into Hi-SOP with the following new contributions:\looseness=-1

1) Conceptually, we introduce a new hierarchical context alignment paradigm that first disentangles geometric and temporal context learning into a local-global hierarchy, and then aggregates them based on semantic consistency for a complementary composition.\looseness=-1

2) Technically, we propose a Geometric Confidence-aware Lifting (GCL) module to explicitly model the geometry with depth distribution confidence for a reliable volumetric feature alignment.
Moreover, the temporal frame features are aligned based on their contextual relevance and ensembled accordingly to achieve mutual compensation. 
To ultimately align the geometric and temporal context in a unified space with a global view, we further design a Depth-Hypothesis-Based Transformation (DHBT) to enable stable geometric-temporal volume composition.\looseness=-1

3) Experimentally, the initial HTCL focuses only on the semantic scene completion task, while Hi-SOP extends the framework for semantic occupancy prediction tasks including semantic scene completion and LiDAR semantic segmentation. The project website is available at \url{https://arlo0o.github.io/hisop.github.io/}.

\section{Related Work}

\subsection{Semantic Occupancy Prediction}

{
Semantic Occupancy Prediction (SOP), also referred to as Semantic Scene Completion (SSC), represents a comprehensive 3D perception task that concurrently tackles semantic segmentation and scene completion~\cite{behley2019semantickitti,cai2021semantic,philion2020lift,li2023stereoscene,li2024hierarchical2}. 
Current studies can be broadly categorized into LiDAR-based and camera-based approaches.
{LiDAR-based methods have seen significant advancements driven by innovative neural architectures and intermediate representations specifically designed to address the sparsity and irregularity of LiDAR data~\cite{xia2023scpnet,zhang2018efficient,rist2021semantic,garbade2019two,cortinhal2020salsanext,zhang2020polarnet,milioto2019rangenet++}. For instance, SCPNet~\cite{xia2023scpnet} tackles challenges such as sparse input and label noise by introducing Multi-Path Blocks for multi-scale feature aggregation and a Dense-to-Sparse Knowledge Distillation strategy to enhance single-frame representation learning. SalsaNext~\cite{cortinhal2020salsanext} improves semantic segmentation of LiDAR point clouds through a residual dilated convolution stack and Bayesian uncertainty estimation, achieving state-of-the-art performance on the SemanticKITTI dataset. PolarNet~\cite{zhang2020polarnet} proposes a polar bird’s-eye-view representation to address the uneven spatial distribution of points, enabling efficient and accurate online semantic segmentation with low latency. Similarly, RangeNet++~\cite{milioto2019rangenet++} leverages range images as an intermediate representation and introduces a novel post-processing algorithm to mitigate discretization errors, achieving real-time, high-accuracy semantic segmentation on embedded hardware.}
Camera-based 3D SOP has recently gained traction due to its cost-effectiveness and portability~\cite{behley2019semantickitti,silberman2012indoor,song2017semantic,straub2019replica,cai2021semantic,rist2021semantic,yan2021sparse,cheng2021s3cnet,wu2020scfusion,roldao2020lmscnet,li2020anisotropic,li2023bevdepth,cao2022monoscene,li2023voxformer,huang2023tri}. MonoScene~\cite{cao2022monoscene} pioneered the inference of geometry and semantics from a single RGB image using 2D-3D feature projection. Following this innovation, numerous studies have expanded the scope of camera-based 3D scene perception~\cite{huang2023tri,zhang2023occformer,wei2023surroundocc,li2023stereoscene}. OccFormer~\cite{zhang2023occformer} employs a monocular depth net and context net to lift voxel feature volume, which is processed with a dual-path transformer block for semantic occupancy prediction.
TPVFormer~\cite{huang2023tri} introduces a tri-perspective view to enhance the detailed representation of a 3D scene. SurroundOcc~\cite{wei2023surroundocc} estimates dense 3D occupancy using multi-view image inputs. VPD~\cite{li2024time} utilizes conditional diffusion models for 3D perception tasks, including multi-view stereo and semantic occupancy prediction.  
In this paper, we advocate for leveraging reliable temporal data to dynamically integrate semantic context and compensate for incomplete observations.}\looseness=-1

\subsection{Geometry Learning in BEV Representation}

The bird's-eye view (BEV) is a prevalent representation in 3D object detection, offering a comprehensive depiction of layouts and strong hallucination capabilities from a top-down perspective~\cite{philion2020lift,huang2021bevdet,li2022bevstereo,li2022bevformer,li2023bev}. The Lift-Splat approach~\cite{philion2020lift} initially introduced the extraction of BEV representations from multiple cameras by implicitly unprojecting 2D visual inputs through estimated depth distributions. To improve geometric modeling in the lifting process, BEVDepth~\cite{li2023bevdepth} employs a camera-aware monocular depth estimation module, enhancing depth accuracy in BEV-based 3D detection. 
As an effective representation for 3D scenarios, BEV representations have also been effectively utilized in recent occupancy-based perception works~\cite{zhang2023occformer, wei2023surroundocc, li2023stereoscene}. Notably, StereoScene~\cite{li2023stereoscene} takes advantage of stereo matching technology and utilizes stereo images to improve the geometric information in the BEV representation and achieves remarkable enhancements. In this study, we develop a new framework that incorporates monocular or stereo depth estimation to explicitly model geometric information with depth distribution confidence awareness.
Moreover, the geometric context is aligned with the temporal context into a unified space through depth-hypothesis-based transformation for stable representation aggregation.\looseness=-1

\subsection{Temporal Modeling in 3D Visual Perception}

The incorporation of temporal information has gained prominence in applications such as temporal 3D object detection~\cite{li2021enforcing,kopf2021robust,liu2023petrv2} and video depth estimation~\cite{chen2024neuralrecon,wang2023crafting}, enhancing overall prediction accuracy. Temporal 3D object detection generally targets coarse-grained, regional-level predictions~\cite{liu2023petrv2,lin2022sparse4d}, whereas video depth estimation techniques strive to establish correspondences across sequential video frames~\cite{long2021multi,cai2023riav}. Nevertheless, such approaches fall short in SOP, where capturing fine-grained features is crucial for dense semantic perception. 
VoxFormer~\cite{li2023voxformer} establishes the first temporal framework for camera-based SOP by merely stacking features from different frames, yet the temporal correspondence modeling for the dense perception task of SOP remains unexplored. 
\emph{GaussianWorld~\cite{zuo2025gaussianworld} proposes to exploit the scene evolution by considering the continuity of 3D occupancy. However, it fuses context from different temporal frames without explicit pattern similarity modeling, which may lead to unstable representation learning due to unaligned semantic content.
GDFusion~\cite{chen2025rethinking} incorporates the formulation of RNNs to implement gradient descent for temporal information integration, considering temporal motion, temporal geometry, and scene-level temporal cues.
Nevertheless, the lack of explicit temporal correspondence modeling may result in unstable learning, and the RNN-based formulation is susceptible to cumulative errors~\cite{li2017diffusion, mao2022review,li2024time}.} 
In contrast, our approach explicitly models temporal context correlation through homography warping and affinity-aware dynamic refinement, enabling reliable aggregation of aligned temporal content while mitigating the impact of incomplete observations.\looseness=-1

\begin{figure*}[!ht]
\vspace{-10pt}
\hsize=\textwidth %
\centering
\includegraphics[width=0.8\textwidth]{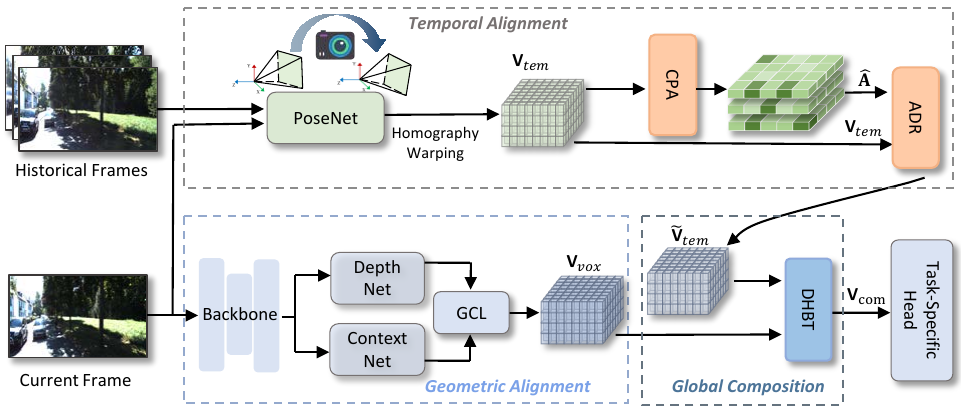}
\vspace{-5pt}
\caption{Overall framework of our proposed hierarchical context alignment scheme, which is composed of the Geometric Alignment, the Temporal Alignment, and the Global Composition. 
The Geometric Confidence-awareness Lifting (GCL) module is introduced to facilitate explicit geometric alignment with depth distribution confidence.
The Cross-frame Pattern Affinity (CPA) measurement and Affinity-based Dynamic Refinement (ADR) module are presented to quantify the regional
contextual relevance and dynamically refine the feature sampling locations based on the relevance information, respectively.
Afterward, the Global Composition with the Depth-Hypothesis-Based Transformation (DHBT) module is introduced to aggregate the disentangled relevant content for reliable fine-grained SOP.\looseness=-1}
\label{figoverall}
 \vspace{-10pt}
\end{figure*}

\section{Methodology}

\subsection{Overview}

\subsubsection{Preliminary}

Given a sequence of temporal RGB images $I_{{set}}^{{rgb}} = \{I_t^{{rgb}}, I_{t-1}^{{rgb}}, \ldots \}$, our objective is to estimate the semantic voxel grid for semantic scene completion or LiDAR segmentation~\cite{cao2022monoscene,zhang2023occformer}. We focus on current and historical image frames, excluding future frames~\cite{li2023voxformer} to devise a practical method for real-world applications. The scene is represented as a voxel grid $\mathbf{V}$ with dimensions $\mathbb{R}^{H \times W \times Z}$, where $H$, $W$, and $Z$ denote the height, width, and depth of the grid, respectively. Each voxel within this grid is classified into one of the semantic categories in the set $\{c_0, c_1, \ldots, c_N\}$, where $c_0$ indicates empty space and $\{c_1, c_2, \ldots, c_N\}$ correspond to $N$ distinct semantic classes. With the proposed framework $\Theta$, we seek to learn a transformation defined as:\looseness=-1

\begin{equation}\label{eq1}
\widehat{\mathbf{V}} = \Theta(I_t^{{rgb}}, I_{t-1}^{{rgb}}, \ldots),
\end{equation}
where $\widehat{\mathbf{V}}$ represents the estimated 3D semantic voxel grid, which aims to approximate the ground truth semantic occupancy or LiDAR semantic labels.
For LiDAR semantic segmentation, we use the LiDAR data only for point query to compute evaluation metrics following previous works~\cite{huang2023tri,zhang2023occformer}.\looseness=-1

\subsubsection{Architectural Design Comparison and Analysis}

To estimate high-quality 3D semantic voxel grid $\widehat{\mathbf{V}}$, existing methods attempt to optimize the scene modeling process from geometric or temporal perspectives while neglecting the misalignment issue. Specifically, the geometric modeling solution~\cite{zhang2023occformer} leverages naive monocular depth estimation for semantic voxel grid estimation and neglects the uncertainty inherent in the depth estimation process:
\begin{equation} 
\widehat{\mathbf{V}} = \texttt{Extr} (I_t^{{rgb}}, I_{t-1}^{{rgb}}, \ldots) \otimes \texttt{Mono} (I_t^{{rgb}}, I_{t-1}^{{rgb}}, \ldots),
\end{equation}
where $\otimes$ denotes the outer product. $\texttt{Extr}$ and $\texttt{Mono}$ represent the 2D feature extractor and monocular depth estimator, respectively.
Such a process inevitably causes geometric ambiguity when the estimated depth is integrated with the corresponding 2D features.
On the other hand, the temporal modeling solution~\cite{li2023voxformer} straightforwardly stacks the temporal image frames to construct the semantic voxel grid $\widehat{\mathbf{V}}$, which ignores the positional changes of shared semantic content across different perspectives:
\begin{equation} 
\widehat{\mathbf{V}} = \texttt{DA} (\texttt{Stack} (I_t^{{rgb}}, I_{t-1}^{{rgb}}, \ldots) ),
\end{equation}
where $\texttt{DA}$ denotes deformable attention. Such a simple straightforward aggregation process could lead to blurred predictive information and compromise the stability of the semantic occupancy prediction.\looseness=-1
 
{To tackle the misalignment issue, we explore first disentangling the complicated semantic scene comprehension into geometric and temporal context alignment, and further align these contexts globally to compose them together to construct the reliable semantic voxel grid $\widehat{\mathbf{V}}$:\looseness=-1
\begin{equation} 
\widehat{\mathbf{V}} = \texttt{Compose} ( \texttt{Geo} ( I_t^{{rgb}}, I_{t-1}^{{rgb}}, \ldots),  \texttt{Tem}( I_t^{{rgb}}, I_{t-1}^{{rgb}}, \ldots) ),
\end{equation}
where $I_t^{{rgb}}$ represents the input temporal RGB image at time step $t$.
The functions $\texttt{Geo}$ and $\texttt{Tem}$ denote geometric context extraction and temporal context extraction, respectively, while $\texttt{Compose}$ represents the global fusion of the two contextual representations.\looseness=-1}
In this way, our methods take advantage of the complementary merits of the geometric and temporal representations in a disentangled local-global architecture, which are hierarchically aligned for a reliable context composition.\looseness=-1

\begin{figure}[!t]
\begin{center} 
\vspace{-0pt}
\includegraphics[width=0.99\linewidth]{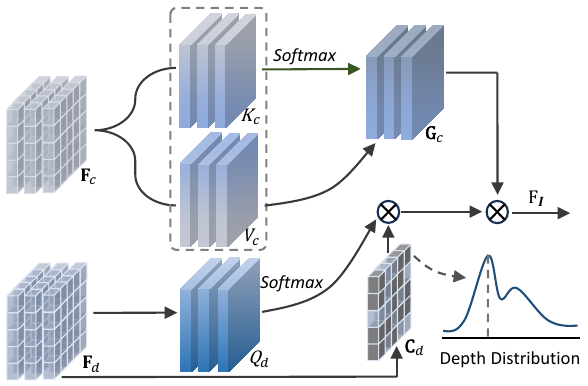}
 \vspace{-5pt}
\caption{The structure of the proposed Geometric Confidence-aware Lifting (GCL) module, which explicitly models the geometric information with
depth distribution confidence.}
\label{module_gcl}
 \vspace{-20pt}
\end{center}
\end{figure}

\subsubsection{Overall Framework}

Specifically, as depicted in Figure~\ref{figoverall}, the overall framework of our proposed method mainly consists of three components: Geometric Alignment in the lower branch, Temporal Alignment in the upper branch, and Reliable Context Aggregation for fine-grained semantic occupancy prediction.\looseness=-1

\noindent\textbf{Geometric Alignment.}
The voxel feature volume $\mathbf{V}_{{vox}}$ is constructed using a UNet architecture based on a pre-trained EfficientNetB7~\cite{tan2019efficientnet}. The network initially generates features with spatial dimensions of $\mathbb{R}^{H/4 \times W/4}$. We then adopt the lifting process following previous studies~\cite{philion2020lift,li2023bevdepth,li2024bridging}, to form $\mathbf{V}_{{vox}}$ from the contextual information and depth distribution. 
For depth distribution modeling, we use off-the-shelf monocular~\cite{bhat2021adabins} or stereo~\cite{cheng2020hierarchical} depth estimation networks. By default, we leverage stereo depth estimation to form our stereo-based pipeline of \emph{Hi-SOP(S)}. Additionally, a monocular-based pipeline of \emph{Hi-SOP(M)} is presented to enhance versatility for scenarios lacking stereo inputs.
To facilitate reliable geometry alignment when constructing $\mathbf{V}_{{vox}}$, we employ a Geometric Confidence-awareness Lifting (GCL) module, which is detailed in Section~\ref{gcl}.\looseness=-1

\noindent\textbf{Temporal Alignment.}
To construct the temporal feature volume $\mathbf{V}_{{tem}}$, we feed current and historical frames into a lightweight PoseNet~\cite{guizilini20203d,cai2023riav} to generate temporal feature volume $\mathbf{V}_{{tem}}$ using homography warping. Different from computing matching costs in typical temporal depth estimation methods~\cite{long2021multi,watson2021temporal,cai2023riav}, we aim to preserve context features within $\mathbf{V}_{{tem}}$. The details are presented in Section~\ref{sec_tvc}.\looseness=-1

Following that, we leverage the temporal volume $\mathbf{V}_{{tem}}$ to generate the cross-frame affinity $\hat{\mathbf{A}}$, quantifying contextual relevance/correspondences between current and historical features. This affinity is then used to reassemble the temporal content and dynamically refine the sampling locations, resulting in a reliable aligned temporal volume $\widetilde{\mathbf{V}}_{{tem}}$. Further details on Cross-frame Pattern Affinity (CPA) and Affinity-based Dynamic Refinement (ADR) are  presented in Sections~\ref{cpa}
and~\ref{adr}, respectively.\looseness=-1

\noindent\textbf{Global Composition.}
To construct a reliable unified representation with semantically {consistency}, the temporal feature volume $\widetilde{\mathbf{V}}_{{tem}}$ is aligned with the voxel feature volume $\mathbf{V}_{{vox}}$ in a global view through depth-hypothesis based transformation.
These two volumes are composed together to generate the composed volume $\mathbf{V}_{{com}}$. The Depth-Hypothesis-Based Transformation (DHBT) module is detailed in Section~\ref{DHBT}.

\subsection{Geometric Alignment with Confidence-aware Lifting}\label{gcl}
As introduced in previous works~\cite{philion2020lift,li2024bridging}, the lifting process establishes volumetric features that store pixel-level context with their associated depth distribution. 
Nonetheless, \cite{li2024bridging} highlights that this process is inherently ambiguous and prone to producing unreliable representations in challenging regions (\textit{e.g.}, severe occlusion and high reflection) where the depth estimation results are unreliable~\cite{li2024bridging}.
To facilitate reliable geometry modeling within the voxel feature volume $\mathbf{V}_{{vox}}$, we propose a Geometric Confidence-aware Lifting (GCL) module to explicitly model the geometric information with depth distribution confidence.\looseness=-1

As shown in Figure~\ref{module_gcl}, the module takes the depth feature $\textbf{F}_{d}$ from the depth net and the context feature $\textbf{F}_{c}$ from the context net as inputs.
To establish pixel-level reliable information for dense prediction, we develop a depth confidence-aware cross-attention mechanism to explicitly indicate the confidence information of the depth distribution and take advantage of the relevant context to complement the low-confidence regions.
Specifically, to project $\textbf{F}_{d}$ to a confidence map $\textbf{C}_d$ 
, we first adopt $softmax$ to convert depth cost value $d_i$ of $\textbf{F}_{d}$ into a probability form, and then take out the highest probability value among all depth hypothesis planes along the depth dimension as the prediction confidence. The process is formally written as:
\begin{equation}
{\textbf{C}_d= \mathrm{WTA}(\phi ( \textbf{F}_{d} )) = \mathrm{WTA} \left\{   \frac{\exp(d_i)} 
 {\sum_{j=1}^{D_{max}}\exp(d_j)}  \right\},}
\end{equation}
where the $softmax$ is applied across the depth dimension and $\mathrm{WTA}$ represents winner-takes-all operation. $D_{max}$ denotes the length of the depth dimension. 

Next, we utilize the depth confidence information to enforce the cross-attention for pixel-level reliable geometric modeling. 
Specifically, we obtain the query $Q_d$ from $\textbf{F}_{d}$  by flattening in spatial and depth dimensions following the standard protocol~\cite{wang2018non,liao2022wt}. 
Similarly, the context feature $\textbf{F}_{c}$ is forwarded and its key and value are denoted as $K_c, V_c$, respectively.
To reduce computational and memory consumption, we follow~\cite{katharopoulos2020transformers,kitaev2020reformer} to compute linear cross-attention:\looseness=-1

\begin{equation} \label{eqca}
\begin{split}
\textbf{F}_{I} &=  Atten(Q_{d},K_{c},V_{c} )  \\
&=  \phi_q(Q_{d}) \odot \textbf{C}_d  (\textbf{G}_{c} ), \\
 &=  \phi_q(Q_{d}) \odot \textbf{C}_d ( \phi_k{ (K_{c}) }^{T} V_{c} ),
\end{split}
\end{equation}
where $\textbf{F}_{I}$ represents the relabel interacted feature, 
$\phi_q$ and $\phi_k$ denote the softmax function along each row and column of the input matrix, respectively. 
$\textbf{G}_{c}$ represents global contextual vectors of $\textbf{F}_{c}$. 
$\odot$ represents the element-wise product, through which the reliable geometry information is preserved while low-confidence information is suppressed.
Finally, the voxel feature volume $\mathbf{V}_{{vox}}$ is obtained from the outer product between the context features $\textbf{F}_{c}$ and the relabel interacted feature $\textbf{F}_{I}$.\looseness=-1

\subsection{Temporal Alignment with Feature Volume Construction}\label{sec_tvc}
The fine-grained nature of the SOP task requires constructing temporally aligned features for accurate and robust perception. Instead of simply stacking input images from various viewpoints as \cite{li2023voxformer}, we propose to align the temporal invariant content using an explicit homography transformation.\looseness=-1

As illustrated in Figure~\ref{figoverall}, we first process the current and historical frames using a lightweight PoseNet~\cite{guizilini20203d,watson2021temporal} to generate the relative camera poses for photometric reprojection. Subsequently, we utilize these frames to generate both the current feature map $F_{t}$ and historical feature maps $\{ F_{t-1}, \cdots, F_{t-n} \}$. Following~\cite{cai2023riav,watson2021temporal}, we construct the warped historical features by applying homography warping using the relative camera poses and alternate depth hypothesis planes, which is defined as:\looseness=-1

\begin{equation}
\mathrm{Warp}(\mathbf{p}) =\mathbf{K}_i \cdot\left(\mathbf{R}_{0, i} \cdot\left(\mathbf{K}_0^{-1} \cdot \mathbf{p} \cdot d_j\right)+\mathbf{t}_{0, i}\right),
\end{equation}
where $\left\{\mathbf{K}_i\right\}_{i=0}^{N-1}$ represent the camera intrinsic parameters and $\left\{\left[\mathbf{R}_{0, i} \mid \mathbf{t}_{0, i}\right]\right\}_{i=1}^{N-1}$ denote the extrinsic parameters, respectively. The variable $d_j$ represents the hypothesized depth for pixel $\mathbf{p}$ in $F_t$. Following this, we aggregate all the warped historical features to create a historical feature volume $\textbf{V}_{tem}^{his}$, which ensures geometric compatibility across varying depth values between the current and historical frames.
Next, we lift $F_t$ along the depth dimension as described in \cite{watson2021temporal,newcombe2011dtam}, generating the current feature volume $\textbf{V}_{tem}^{cur}$. By concatenating $\textbf{V}_{tem}^{his}$ with $\textbf{V}_{tem}^{cur}$ following \cite{watson2021temporal}, we construct the composite temporal feature volume $\textbf{V}_{tem}$:\looseness=-1
\begin{equation}
\begin{aligned}
\textbf{V}_{tem} &= \texttt{Concat} \left \{  (\textbf{V}_{tem}^{cur}, \textbf{V}_{tem}^{his}), \mathrm{dim}=\mathbb{C} \right \} \\
&=  \texttt{Concat} \left \{  \texttt{Lift}(F_t), \texttt{Warp}(F_{t-1}, \cdots, F_{t-n} ) \right \}.   \\
\end{aligned}
\end{equation}

The temporal feature volume $\textbf{V}_{tem}$ enhances semantic scene modeling by aligning contextual features across different time steps. In Section~\ref{cpa} and Section~\ref{adr}, we will detail the methodology for harnessing reliable information through contextual correspondence within $\textbf{V}_{tem}$.\looseness=-1

\noindent\textbf{Why Feature Volume Instead of Cost Volume?}
Conventional temporal depth estimation networks typically build cost volumes by computing the Hadamard product~\cite{long2021multi,cai2023riav} or absolute differences~\cite{watson2021temporal} between different feature maps:
\begin{equation}    
\textbf{C(d)}=\frac{1}{N} \sum_{i=1}^{N} \texttt{Match} ( {f_0^{ref} , \tilde{f}_i^{warp} }  ),
\end{equation}
where $\textbf{C}$ denotes the constructed cost volume with depth hypothesis $d$. $\texttt{Match}$ represents the matching operation. $f_0^{ref}$ and $\tilde{f}_i^{warp}$ denotes the reference feature and warped feature from $i^{th}$ image frame.
In contrast, our approach centers around the construction of feature volumes, with the objective of more effectively preserving the extensive context crucial for the Semantic Occupancy Prediction (SOP) task. The biggest difference between these two tasks stems from the nature of camera-based SOP. As illustrated in Equation~\ref{eq1}, SOP is inherently a task for dense perception and reconstruction, rather than a matching problem. Therefore, our method focuses on feature volume instead of cost volume, which preserves the integrity of fine-grained feature context rather than calculating matching costs within the temporal feature volume. Furthermore, to assess the significance of regional patterns within the temporal data, we establish an auxiliary pattern affinity metric between the current and historical features.\looseness=-1

\begin{figure}[!t]
\begin{center} 
\vspace{-0pt}
\includegraphics[width=0.99\linewidth]{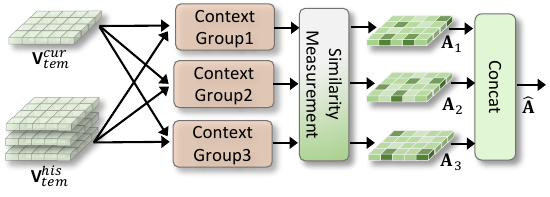}
 \vspace{-10pt}
\caption{The structure of the proposed Cross-frame Pattern Affinity (CPA) measurement, which is proposed to quantify the regional contextual correspondence within the temporal feature volume.
}
\label{module_cpa}
 \vspace{-15pt}
\end{center}
\end{figure}

\begin{figure}[!ht]
\vspace{-0pt}
\centering
\includegraphics[width=0.95\linewidth]{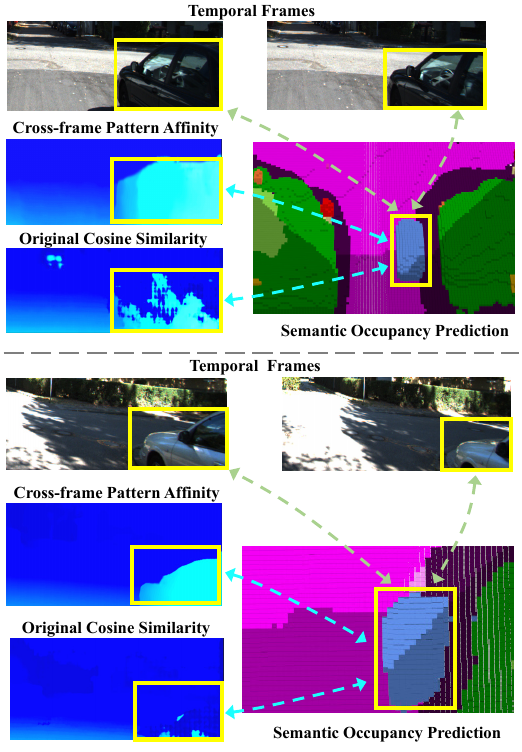}  
\vspace{-5pt}
\caption{Visualization of the heat maps from our proposed Cross-frame Pattern Affinity (CPA) and the original cosine similarity. 
} 
\label{fig_aff}
\vspace{-15pt}
\end{figure} 

\subsection{Cross-frame Pattern Affinity for Relevance Modeling}\label{cpa}

Despite the explicit alignment of the temporal volume, it integrates redundant contexts from various frames, which are inadequate for directly modeling scene representations corresponding to the current frame. Consequently, we introduce the Cross-frame Pattern Affinity (CPA) to quantify the regional contextual relevance between the historical feature volume $\textbf{V}_{his}$ and the current feature volume $\textbf{V}_{cur}$.\looseness=-1

\noindent\textbf{Similarity Measurement Optimization Analysis}.
As a widely applied metric in semantic analysis~\cite{evangelopoulos2013latent, ramachandran2011automated, evangelopoulos2012latent} and information retrieval~\cite{rahutomo2012semantic, korenius2007principal}, Cosine similarity measures correlations effectively. The cosine similarity between two vectors $\alpha$ and $\beta$ is computed as:\looseness=-1
\begin{equation}
\label{cos}
\texttt{sim}(\alpha, \beta) = \texttt{cos}(\vec{\alpha}, \vec{\beta}) = \frac{\vec{\alpha} \cdot \vec{\beta}}{\|\vec{\alpha}\| \ast \|\vec{\beta}\|}.
\end{equation}

Nevertheless, traditional cosine similarity can yield high similarity scores for vectors that are not truly similar~\cite{sarwar2001item}. This issue is acknowledged and addressed through scale-aware isolation~\cite{anastasiu2014l2ap}, which adjusts for variations in pattern scales. However, such methods primarily focus on vector orientations and struggle to assess similarity within densely distributed datasets. To mitigate these limitations, ensemble learning techniques~\cite{xia2015learning}, which utilize a diverse array of independent learners, have been employed to improve the accuracy of similarity assessments in dense environments.\looseness=-1

Given these concerns, we establish the criteria for an optimal similarity measurement strategy in SOP: \emph{incorporation of diverse independent learning} and \emph{scale-aware isolation}. To achieve this, we propose to employ scale-aware isolated cosine similarity and the integration of multi-group context as inputs for affinity computation in dense distributions. Our approach is implemented through two principal steps:\looseness=-1

\begin{itemize}
    \item Incorporation of various pattern scales from multi-group contexts, fostering diverse independent similarity learning for fine-grained SOP.\looseness=-1
    \item Calculation of cosine similarities using scale-aware isolation, followed by their aggregation to ensure accurate pattern affinity measurement.\looseness=-1
\end{itemize}

\begin{figure}[!t]
\begin{center} 
\vspace{-0pt}
\includegraphics[width=0.99\linewidth]{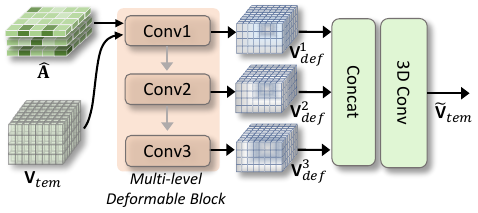}
 \vspace{-15pt}
\caption{The structure of the proposed Affinity-based Dynamic Refinement (ADR) module, which dynamically refines the feature sampling locations based on the identified high-affinity locations and their neighboring relevant regions. 
}
\label{module_adr}
 \vspace{-15pt}
\end{center}
\end{figure}

\noindent\textbf{Multi-group Context Generation.}  
To support the learning of diverse and independent similarities, 3D atrous convolutions with different dilation rates are utilized to develop multi-group contextual features. Specifically, the historical feature volume $\textbf{V}_{tem}^{his}$ undergoes processing through a series of atrous convolutions to produce the historical multi-group context $\textbf{H}_i$ for $i \in \{1, 2, 3\}$, defined as:\looseness=-1

\begin{equation}
\textbf{H}_i =  \texttt{GN} \left ( \delta \left (  \texttt{Atrous}_i (\textbf{V}_{tem}^{his}) \right ) \right ),
\end{equation}
where $\texttt{GN}$ represents group normalization and $\delta$ signifies the GELU activation function. The atrous convolutions are applied in parallel, with dilation rates of 1, 2, and 4. In a similar approach, the current multi-group context $\textbf{C}_i$ is derived symmetrically from the current feature volume $\textbf{V}_{tem}^{cur}$ as follows:\looseness=-1

\begin{equation}
\textbf{C}_i =  \texttt{GN} \left ( \delta \left (  \texttt{Atrous}_i (\textbf{V}_{tem}^{cur}) \right ) \right ).
\end{equation}

\noindent\textbf{Measuring Pattern Affinity for Dense SOP.}
We refined Equation~\ref{cos} with two key modifications to enhance the measurement of pattern affinity, facilitating fine-grained contextual correspondence modeling in SOP. Firstly, we address the multi-scale nature of the group context by computing the pattern affinity $\textbf{A}_{i}$ for each scale $i$. These independent group-scale affinity matrices are then aggregated along the channel dimension.
Secondly, during the affinity calculation for each scale, we adjust for scale variability by subtracting the average values within each group scale, thereby achieving scale-aware isolation. The mathematical representation is as follows:\looseness=-1

\begin{align}
\textbf{A}_{i}  &=  \texttt{sim} ( \textbf{C}_{i}, \textbf{H}_{i}) \\ \notag
 &=  \frac{ \sum_{j=0}^{C} ( \textbf{C}_{i}^{j} - \overline{ \textbf{C}}_{i} )(\textbf{H}^{j}_{i} - \overline{ \textbf{H}}_{i})}{ \sqrt{ {\sum_{j=0}^{C} ( \textbf{C}^{j}_{i} - \overline{ \textbf{C}}_{i} )}^{2} } \sqrt{ {\sum_{j=0}^{C} ( \textbf{H}^{j}_{i} - \overline{ \textbf{H}}_{i} )}^{2} } } , \\ 
 \hat{\textbf{A}} &=  \texttt{Concat}  \left \{ (\textbf{A}_1,\textbf{A}_2,\textbf{A}_3), \mathrm{dim}=\mathbb{C}  \right \} , 
\end{align}
where the affinity matrices $\textbf{A}_{i}$ of different group scales are concatenated along the channel dimension to derive the composite cross-frame pattern affinity $\hat{\textbf{A}}$. The input context matrices $\textbf{C}_{i}$ and $\textbf{H}_{i}$ are considered as high-dimensional vectors across various group scales. The matrices $\overline{ \textbf{C}}_{i}$ and $\overline{ \textbf{H}}_{i}$ denote the averaged context matrices for each respective group scale. As depicted in Figure~\ref{fig_aff}, the Cross-frame Pattern Affinity (CPA) effectively highlights the contextual correspondence within the temporal content.

\begin{figure}[!t]
\begin{center} 
\vspace{-0pt}
\includegraphics[width=0.99\linewidth]{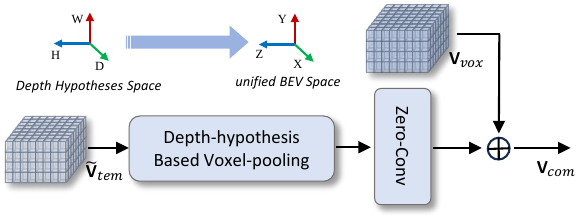}
 \vspace{-10pt}
\caption{The structure of the Depth-Hypothesis-Based Transformation (DHBT), which is proposed to facilitate reliable global composition of the feature volumes.}
\label{module_dhbt}
 \vspace{-15pt}
\end{center}
\end{figure}

\subsection{Affinity-based Dynamic Refinement}\label{adr}

Given our objective of completing and comprehending the 3D scene corresponding to the current frame, it is essential to assign greater weights to the most relevant locations.
Concurrently, exploring their neighboring relevant context is also critical to compensate for incomplete observations.\looseness=-1

To this end, we propose to dynamically refine the feature sampling locations based on the identified high-affinity locations and their neighboring relevant regions.
The above ideas are implemented using 3D deformable convolutions~\cite{dai2017deformable,wang2021patchmatchnet}.
Specifically, dynamic refinement is achieved through the introduction of affinity-based correspondence weights and deformable positional offsets.
In the context of a sampling grid window $K_w$, the formula is expressed as:\looseness=-1

\begin{equation} \label{eqca2}
\begin{split}
\textbf{V}_{def} = \sum_{k=1}^{K_w} w_k \cdot   \textbf{V}_{tem}(\textbf{p}+\textbf{p}_k+\Delta \textbf{p}_k) \cdot a_{k},
\end{split}
\end{equation}
where $K_w$ represents the number of points in the sampling process. $\Delta \textbf{p}_k$ denotes the additional offset in the sampling grid. $w_k$ denotes the spatial feature weight and $a_{k}$ represents the affinity weight from the cross-frame pattern affinity $\hat{\textbf{A}}$.\looseness=-1

To enhance dynamic modeling through hierarchical context, we refine the process by incorporating contextual information across different feature levels. As depicted in Figure~\ref{module_adr}, a multi-level deformable block is constructed, consisting of three cascaded 3D deformable convolutions. These output features are aggregated to form a reliable temporal volume, $\widetilde{\textbf{V}}_{tem}$, as expressed in the following equation:\looseness=-1

\begin{equation} \label{eqca3}
\widetilde{\textbf{V}}_{tem} = \textbf{W} \left( \texttt{Concat} \left\{ (\textbf{V}_{def}^{1}, \textbf{V}_{def}^{2}, \textbf{V}_{def}^{3}), \mathrm{dim}=\mathbb{C} \right\} \right),
\end{equation}
where the multi-level deformable temporal volumes, $\textbf{V}_{def}^{i}$ (where $i \in \{1,2,3\}$), are concatenated along the channel dimension. Subsequently, they are processed using a 3D convolution layer, $\textbf{W}$, to reduce dimensionality.\looseness=-1

\subsection{Global Alignment with Unified Transformation}\label{DHBT}

To globally align the geometric context with the temporal context within a unified space for semantic-consistent aggregation, we present the Depth-Hypothesis-Based Transformation (DHBT) as follows.
Firstly, the temporal volume $\widetilde{\mathbf{V}}_{{tem}}$ is aligned with the voxel feature volume $\mathbf{V}_{{vox}}$ through depth distribution hypothesis. As depicted in Figure~\ref{module_dhbt}, we take the depth hypothesis of $\widetilde{\mathbf{V}}_{{tem}}$ as the distance axis and employ a voxel-pooling operation following~\cite{philion2020lift,li2024bridging} to splat the volumetric features into the unified space.
Following that, we aggregate the voxel feature volume $\mathbf{V}_{{vox}}$ and the temporal feature volume $\widetilde{\mathbf{V}}_{{tem}}$ for reliable information interaction. 
In the initial stages of training, unregulated temporal information may compromise the learning of voxel features. To address this, we employ a flexible element-wise aggregation strategy:

\begin{equation}
\mathbf{V}_{{com}} = \texttt{Zero\_Conv} (\texttt{Voxel\_Pool}(\widetilde{\mathbf{V}}_{{tem}}) ) + \mathbf{V}_{{vox}},
\end{equation}
where $\texttt{Zero\_Conv}$ is the zero convolution as ControlNet~\cite{zhang2023adding} to retain the inherent capabilities of the temporal feature volume. The composed volume $\mathbf{V}_{{com}}$ is then processed through a task-specific head to generate the semantic occupancy voxel or LiDAR semantic labels following previous works~\cite{huang2023tri,zhang2023occformer}.

\noindent\textbf{Why Global Alignment Necessary?}
The Lift-Splat-Shoot (LSS) paradigm is widely employed in bird's-eye view (BEV) representations, which typically aggregates multi-view image features into a unified space based on depth distributions, enabling the transformation of 2D images into 3D representations~\cite{huang2021bevdet,li2022bevformer,zhang2023occformer}. The transformation is performed via the outer product between the 2D image feature of $i^th$ frame and its corresponding depth distribution: $\textbf{F}_i^{BEV}=f_i^{2d} \otimes d_i^{dis}$.
The core motivation behind such an operation lies in the necessity for constructing a uniform contextual distribution that facilitates more reliable representation and stable learning processes. However, in our framework, 
the temporal volume $\widetilde{\mathbf{V}}_{{tem}}$ and the voxel feature volume $\mathbf{V}_{{vox}}$ 
are initially misaligned due to different construction strategies used in their respective representation spaces.  Therefore, we employ a unified global transformation approach with the depth distribution hypothesis, to construct reliable volumetric representations.

\subsection{Training Objectives} 
We follow the basic learning objective of MonoScene~\cite{cao2022monoscene} for semantic occupancy prediction. 
Standard semantic loss $\mathcal{L}_{{sem}}$ and geometry loss $\mathcal{L}_{{geo}}$ are leveraged for semantic and geometry supervision, while an extra class weighting loss $\mathcal{L}_{ce}$ is also added.
To further enforce the ensembled volume, we adopt a binary cross entropy loss $\mathcal{L}_{depth}$ to encourage the sparse depth distribution. The overall learning objective of this framework is formulated as follows:\looseness=-1
\begin{equation}
   { \mathcal{L} = \mathcal{L}_{depth} + \mathcal{\lambda}_{ce} \mathcal{L}_{ce}. }
\end{equation}
where several $\lambda$s are balancing coefficients.

\begin{table*}[htbp]
\vspace{-10pt}
\begin{center}
\scriptsize
\caption{\textbf{Quantitative comparison} with the state-of-the-art camera-based {semantic occupancy prediction} methods on the SemanticKITTI validation set. 
The {``S-T'', ``S''} and {``M''} denote temporal stereo images, single-frame stereo images, and single-frame monocular images, respectively. The top two performers are marked \textbf{bold} and \underline{underline}.
}
\vspace{-0pt}
\renewcommand\tabcolsep{8.0pt}
\resizebox{0.98\textwidth}{!}{%
\begin{tabular}{l|cccc|ccc}
\toprule
{Methods}  &\cellcolor[gray]{0.93}Hi-SOP(S) & HTCL-S  & VoxFormer-T  & VoxFormer-S & OccFormer & TPVFormer  & MonoScene  \\ \midrule

{Input}    & \cellcolor[gray]{0.93}{S-T} & {S-T}  &{S-T} &{S} &{M}  &{M}  &{M}  \\ \midrule
\textbf{IoU }  & \cellcolor[gray]{0.93}\textbf{45.56} & \underline{45.51}   & 44.15 &44.02 & 36.50 & 36.61&37.12 \\ \midrule
\textbf{mIoU }   & \cellcolor[gray]{0.93}\textbf{18.19}  & \underline{17.13}  &13.35 &12.35 & 13.46&11.36&11.50   \\ \midrule

\crule[carcolor]{0.13cm}{0.13cm} {car} & \cellcolor[gray]{0.93}\underline{34.07} & \textbf{34.30}   & 26.54 & 25.79 &25.09& 23.81 & 23.55 \\

\crule[bicyclecolor]{0.13cm}{0.13cm} {bicycle} & \cellcolor[gray]{0.93}\textbf{4.42} & \underline{3.99}  & 1.28 & 0.59 &0.81 & 0.36 &0.20 \\

\crule[motorcyclecolor]{0.13cm}{0.13cm} {motorcycle} & \cellcolor[gray]{0.93}\textbf{3.96} & \underline{2.80} & 0.56 & 0.51 &1.19 &0.05 &0.77  \\

\crule[truckcolor]{0.13cm}{0.13cm} {truck} & \cellcolor[gray]{0.93}\textbf{25.25} & \underline{20.72}& 8.10 & 7.26 &25.53 &8.08 & 7.83     \\

\crule[othervehiclecolor]{0.13cm}{0.13cm} {other-veh.} & \cellcolor[gray]{0.93}\textbf{16.96} & \underline{11.99}  &7.81 &3.77&8.52 &4.35 &3.59  \\

\crule[personcolor]{0.13cm}{0.13cm} {person} & \cellcolor[gray]{0.93}\textbf{3.36} & \underline{2.56} &1.93 &1.78 &2.78 &0.51  &1.79  \\

\crule[bicyclistcolor]{0.13cm}{0.13cm} {bicyclist} & \cellcolor[gray]{0.93}\textbf{6.48} & {2.30} &1.97 &\underline{3.32} &2.82 &0.89 &1.03   \\

\crule[motorcyclistcolor]{0.13cm}{0.13cm} {motorcyclist} & \cellcolor[gray]{0.93}0.00 & {0.00} &{0.00}  &0.00 &{0.00}  &0.00&{0.00}  \\

\crule[roadcolor]{0.13cm}{0.13cm} {road} & \cellcolor[gray]{0.93}\textbf{63.86} & \underline{63.70}  &53.57 &54.76 &58.85 &56.50 &57.47    \\

\crule[parkingcolor]{0.13cm}{0.13cm} {parking} & \cellcolor[gray]{0.93}\textbf{25.94} & \underline{23.27}  &{19.69} &15.50 &19.61 &20.60 &15.72  \\

\crule[sidewalkcolor]{0.13cm}{0.13cm} {sidewalk}  & \cellcolor[gray]{0.93}\textbf{32.71}  & \underline{32.48}  &26.52 &26.35&26.88 &25.87 &27.05    \\

\crule[othergroundcolor]{0.13cm}{0.13cm} {other.grd} & \cellcolor[gray]{0.93}\textbf{1.18} & {0.14}&0.42 &0.70 &19.61 &20.60 &\underline{0.87}    \\

\crule[buildingcolor]{0.13cm}{0.13cm} {building} & \cellcolor[gray]{0.93}\textbf{24.56}  &\underline{24.13} & {19.54}  &17.65 &14.40 &13.88 &14.24   \\

\crule[fencecolor]{0.13cm}{0.13cm} {fence} & \cellcolor[gray]{0.93}\underline{9.30} & \textbf{11.22}  &{7.31} &7.64 &5.61 &5.94 &6.39  \\

\crule[vegetationcolor]{0.13cm}{0.13cm} {vegetation} & \cellcolor[gray]{0.93}\underline{26.61} & \textbf{26.96}  &26.10 &24.39 &19.63&16.92  &18.12  \\

\crule[trunkcolor]{0.13cm}{0.13cm} {trunk} & \cellcolor[gray]{0.93}\textbf{9.92} & \underline{8.79}  &6.10 &5.08 &3.93 &2.26 &2.57  \\

\crule[terraincolor]{0.13cm}{0.13cm} {terrain} & \cellcolor[gray]{0.93}\textbf{38.89} & \underline{37.73}  &33.06 &29.96 &32.62 &30.38 &30.76   \\

\crule[polecolor]{0.13cm}{0.13cm} {pole} & \cellcolor[gray]{0.93}\underline{11.41} & \textbf{11.49}  &9.15 &7.11 &4.26 &3.14 &4.11  \\

\crule[trafficsigncolor]{0.13cm}{0.13cm} {traf.sign} & \cellcolor[gray]{0.93}\underline{6.70} & \textbf{6.95}   &4.94 &4.18 &2.86 &1.52 &2.48 \\  \bottomrule

\end{tabular}
}
\label{tabq1}        
\vspace{-20pt}
\end{center}
\end{table*}

\begin{table*}[htbp]
\vspace{-10pt}
\begin{center}
\scriptsize
\caption{\textbf{Quantitative results} with the state-of-the-art {semantic occupancy prediction} methods on the SemanticKITTI test set. 
The {``S-T'', ``S''} and {``M''} denote temporal stereo images, single-frame stereo images, and single-frame monocular images, respectively. The top two performers are marked \textbf{bold} and \underline{underline}.\looseness=-1
}
\vspace{-0pt}
\renewcommand\tabcolsep{6.7pt}
\resizebox{0.99\textwidth}{!}{%
\begin{tabular}{l|cccc|cccc}
\toprule
{Methods}  &\cellcolor[gray]{0.93}Hi-SOP(S) & HTCL-S    & VoxFormer-T  & VoxFormer-S  & OccFormer &SurroundOcc & TPVFormer  & MonoScene  \\ \midrule

{Input}    & \cellcolor[gray]{0.93}{S-T}  & {S-T}  &{S-T} &{S} &{M} &{M} &{M} & {M}  \\ \midrule
\textbf{IoU } & \cellcolor[gray]{0.93}\textbf{44.57} & \underline{44.23}   & {43.21} &42.95 & 34.53  & 34.72 &34.25 &34.16      \\ \midrule
\textbf{mIoU }  &  \cellcolor[gray]{0.93}\textbf{17.49}  & \underline{17.09}  & {13.41}  &12.20 &12.32 & 11.86 &11.26   &11.08      \\ \midrule

\crule[carcolor]{0.13cm}{0.13cm} {car} & \cellcolor[gray]{0.93}\textbf{27.35} & \underline{27.30}  & {21.70} & 20.80 & 21.60 &20.60 & 19.20   &  18.80  \\

\crule[bicyclecolor]{0.13cm}{0.13cm} {bicycle} & \cellcolor[gray]{0.93}\textbf{2.99} & {1.80} &\underline{1.90} & 1.00& 1.50 &1.60 & 1.00 & {0.50} \\

\crule[motorcyclecolor]{0.13cm}{0.13cm} {motorcycle} & \cellcolor[gray]{0.93}\textbf{2.59}  & \underline{2.20} &{1.60} &0.70 & {1.70} &1.20 & 0.50  &  {0.70}   \\

\crule[truckcolor]{0.13cm}{0.13cm} {truck} &\cellcolor[gray]{0.93}\textbf{7.18}  & \underline{5.70} & {3.60} &3.50 & {1.20} &1.40  & {3.70}  &  {3.30}   \\

\crule[othervehiclecolor]{0.13cm}{0.13cm} {other-veh.} & \cellcolor[gray]{0.93}\textbf{7.19} & \underline{5.40}  &4.10 & 3.70 & 3.20 &{4.40} & 2.30  &{4.40}   \\

\crule[personcolor]{0.13cm}{0.13cm} {person} & \cellcolor[gray]{0.93}\underline{1.68} & {1.10} &{1.60} &1.40& \textbf{2.20} &1.40 &{1.10}  & {1.00} \\

\crule[bicyclistcolor]{0.13cm}{0.13cm} {bicyclist} & \cellcolor[gray]{0.93}\textbf{4.81} & \underline{3.10}  & 1.10 & {2.60} &1.10 &2.00  & {2.40} & {1.40}   \\

\crule[motorcyclistcolor]{0.13cm}{0.13cm} {motorcyclist} & \cellcolor[gray]{0.93}\textbf{1.06} & \underline{0.90}  & 0.00 &0.20 & 0.20 &0.10 & 0.30   &  {0.40}  \\

\crule[roadcolor]{0.13cm}{0.13cm} {road} & \cellcolor[gray]{0.93}\underline{63.95} & \textbf{64.40}   &54.10 &53.90 & 55.90 &{56.90} &{55.10}    &{54.70}  \\

\crule[parkingcolor]{0.13cm}{0.13cm} {parking} &\cellcolor[gray]{0.93}\textbf{35.58} & \underline{33.80} & 25.10& 21.10 &{31.50} &30.20 & {27.40}   & {24.80}  \\

\crule[sidewalkcolor]{0.13cm}{0.13cm} {sidewalk}   &\cellcolor[gray]{0.93}\underline{34.27}  & \textbf{34.80}  &26.90 &25.30 & {30.30} &28.30 & {27.20} & {27.10}   \\

\crule[othergroundcolor]{0.13cm}{0.13cm} {other.grd} &\cellcolor[gray]{0.93}\textbf{13.77} & \underline{12.40}  & {7.30} &5.60 &6.50 &6.80 & 6.50 & {5.70}    \\

\crule[buildingcolor]{0.13cm}{0.13cm} {building}  & \cellcolor[gray]{0.93}\textbf{25.91} & \underline{25.90}  &{23.50} & 19.80 &15.70 &15.20 & 14.80 & {14.40}  \\

\crule[fencecolor]{0.13cm}{0.13cm} {fence} & \cellcolor[gray]{0.93}\underline{20.15} & \textbf{21.10}  &{13.10} & 11.10 & 11.90 & 11.30 & {11.00} &  {11.10} \\

\crule[vegetationcolor]{0.13cm}{0.13cm} {vegetation} &\cellcolor[gray]{0.93}\textbf{26.07}  & \underline{25.30} & {24.40} & 22.40 & 16.80 &14.90 & 13.90 & {14.90}  \\

\crule[trunkcolor]{0.13cm}{0.13cm} {trunk} & \cellcolor[gray]{0.93}\underline{10.35} & \textbf{10.80} & {8.10} & 7.50 & 3.90 & 3.40 & 2.60   &2.40\\

\crule[terraincolor]{0.13cm}{0.13cm} {terrain} &\cellcolor[gray]{0.93}\underline{30.77}  & \textbf{31.20} & {24.20} &21.30 & 21.30 &19.30  &20.40 & {19.50}   \\

\crule[polecolor]{0.13cm}{0.13cm} {pole} & \cellcolor[gray]{0.93}\underline{8.70} & \textbf{9.00}   & {6.60} & 5.10 & 3.80 &3.90 & 2.90 &3.30   \\

\crule[trafficsigncolor]{0.13cm}{0.13cm} {traf.sign} & \cellcolor[gray]{0.93}\underline{7.90} & \textbf{8.30}  &{5.70} & 4.90 & 3.70 &2.40 & 1.50 &2.10  \\  \bottomrule

\end{tabular}
}
\label{tabq1_}        
\vspace{-0pt}
\end{center}
\end{table*}

\begin{table*}[!ht]
\vspace{-0pt}
\begin{center}
\scriptsize
\caption{\textbf{Quantitative comparison} with the state-of-the-art {semantic occupancy prediction} methods on the NuScenes-Occupancy validation set. 
The top two performers are marked \textbf{bold} and \underline{underline}.
The ``L'', ``M'', ``M-D'' and ``M-T'' denote LiDAR inputs, monocular images, monocular images with depth maps and temporal monocular images, respectively. The LiDAR points are projected and densified to generate the depth maps.\looseness=-1}
\vspace{-0pt}
\renewcommand\tabcolsep{6.7pt}
\resizebox{0.99\textwidth}{!}{%
\begin{tabular}{l|cc|cccc|cc}
\toprule
{Methods}  & \cellcolor[gray]{0.93}Hi-SOP(M) & HTCL-M & JS3C-Net & LMSCNet  & 3DSketch & AICNet~\cite{li2021anisotropic} & TPVFormer  & MonoScene  \\ \midrule

{Input}   & \cellcolor[gray]{0.93}{M-T}   & {M-T}  &{L} &{L} &{M-D} &{M-D} &{M} & {M}  \\ \midrule
\textbf{IoU}& \cellcolor[gray]{0.93}\underline{24.5}  & 21.4 &\textbf{30.2} & 27.3& 25.6& 23.8 & 15.3 & 18.4  \\ \midrule
\textbf{mIoU} & \cellcolor[gray]{0.93}\textbf{16.4} &\underline{14.1} & 12.5  & 11.5& 10.7 & 10.6 & 7.8 & 6.9 \\ \midrule

\rotatebox{0}{\textcolor{barrier1}{$\blacksquare$} {barrier}} & \cellcolor[gray]{0.93}\textbf{15.7} & \underline{14.8} & 14.2 & 12.4 & 12.0& 11.5 & 9.3 & 7.1  \\

\rotatebox{0}{\textcolor{bicycle1}{$\blacksquare$} {bicycle}} & \cellcolor[gray]{0.93}\underline{6.4} &\textbf{10.2} & 3.4& 4.2 &5.1 & 4.0 & 4.1 & 3.9\\

\rotatebox{0}{\textcolor{bus1}{$\blacksquare$} {bus}} & \cellcolor[gray]{0.93}\textbf{15.0} &\underline{14.8} & 13.6 & 12.8 &10.7& 11.8&11.3 &9.3\\

\rotatebox{0}{\textcolor{car1}{$\blacksquare$} {car}} &\cellcolor[gray]{0.93}\textbf{20.6} &\underline{18.9} & 12.0& 12.1 &12.4 & 12.3 &10.1 &7.2\\

\rotatebox{0}{\textcolor{const. veh.1}{$\blacksquare$} {const. veh.}}  & \cellcolor[gray]{0.93}\textbf{12.0} &\underline{7.6} & 7.2& 6.2 &6.5 &5.1 &5.2 &5.6\\

\rotatebox{0}{\textcolor{motorcycle1}{$\blacksquare$} {motorcycle}} &\cellcolor[gray]{0.93}\underline{7.0} &\textbf{11.3}  &4.3 &4.7 &4.0& 3.8 & 4.3 &3.0\\

\rotatebox{0}{\textcolor{pedestrian1}{$\blacksquare$} {pedestrian}} &\cellcolor[gray]{0.93}\underline{11.5} &\textbf{12.3}& 7.3& 6.2 &5.0& 6.2 &5.9  &5.9\\

\rotatebox{0}{\textcolor{traffic cone1}{$\blacksquare$} {traffic cone}} &\cellcolor[gray]{0.93}\underline{7.0} &\textbf{9.6}& 6.8& 6.3 &6.3& 6.0 & 5.3 & 4.4\\

\rotatebox{0}{\textcolor{trailer1}{$\blacksquare$} {trailer}} &\cellcolor[gray]{0.93}7.2 & 5.5 &\textbf{9.2}& \underline{8.8} & 8.0 & 8.2 &  6.8 & 4.9\\

\rotatebox{0}{\textcolor{truck1}{$\blacksquare$} {truck}} &\cellcolor[gray]{0.93}\textbf{14.2} &\underline{13.5}& 9.1& 7.2&7.2 &7.5 & 6.5 & 4.2\\

\rotatebox{0}{\textcolor{drive. suf.1}{$\blacksquare$} {drive. suf.}} &\cellcolor[gray]{0.93}\textbf{46.2} &\underline{32.5} & 27.9& 24.2&21.8 &24.1 & 13.6 & 14.9 \\

\rotatebox{0}{\textcolor{other flat1}{$\blacksquare$} {other flat}} &\cellcolor[gray]{0.93}\textbf{29.5} &\underline{21.7} & 15.3& 12.3 &14.8 & 13.0 & 9.0 & 6.3 \\

\rotatebox{0}{\textcolor{sidewalk1}{$\blacksquare$} {sidewalk}} &\cellcolor[gray]{0.93}\textbf{29.2} &\underline{20.7} & 14.9& 16.6 &13.0 & 12.8& 8.3 & 7.9 \\

\rotatebox{0}{\textcolor{terrain1}{$\blacksquare$} {terrain}} &\cellcolor[gray]{0.93}\textbf{25.2} &\underline{17.7} & 16.2& 14.1 &11.8 & 11.5& 8.0 & 7.4 \\

\rotatebox{0}{\textcolor{manmade1}{$\blacksquare$} {manmade}} &\cellcolor[gray]{0.93}5.00 & 5.8 & \textbf{14.}& \underline{13.9} & 12.0 & 11.6& 9.2 & 10.0 \\

\rotatebox{0}{\textcolor{vegetation1}{$\blacksquare$} {vegetation}} &\cellcolor[gray]{0.93}10.4 & 8.5 & \textbf{24.9}& \underline{22.2} & 21.2 & 20.2& 8.2 & 7.6\\
 \bottomrule
\end{tabular}
}
\vspace{-0pt}
\label{tab_open}
\end{center}
\end{table*}

\begin{table*}[!ht]
\vspace{-0pt}
\begin{center}
\scriptsize
\caption{\textbf{Quantitative comparison} with the state-of-the-art LiDAR semantic segmentation methods on the NuScenes validation set. 
The top two performers are marked \textbf{bold} and \underline{underline}.
The ``L'', ``M'' and ``M-T'' denote LiDAR inputs, monocular images and temporal monocular images, respectively.\looseness=-1 }
\vspace{-0pt}
\renewcommand\tabcolsep{12.0pt}
\resizebox{0.99\textwidth}{!}{
\begin{tabular}{l|ccc|ccc}
\toprule
{Methods}  & \cellcolor[gray]{0.93}Hi-SOP(M)  & OccFormer & TPVFormer  & SalsaNext & PolarNet  & RangeNet++  \\ \midrule

{Input}   & \cellcolor[gray]{0.93}{M-T}   & {M}  &{M} &{L} &{L} &{L}   \\ \midrule
\textbf{mIoU} & \cellcolor[gray]{0.93}\textbf{73.7} &{68.1} & 59.3 & \underline{72.2} & 71.0 & 65.5 \\ \midrule

\rotatebox{0}{\textcolor{nbarrier}{$\blacksquare$} barrier} & \cellcolor[gray]{0.93}{71.5} & {69.2} & 64.9  & \textbf{74.8} & \underline{74.7} & 66.0 \\

\rotatebox{0}{\textcolor{nbicycle}{$\blacksquare$} bicycle} & \cellcolor[gray]{0.93}\textbf{43.8} & {36.9} &{27.0} & \underline{34.1} &28.2 & 21.3 \\

\rotatebox{0}{\textcolor{nbus}{$\blacksquare$} bus} & \cellcolor[gray]{0.93}\textbf{92.5} &\underline{91.2} & 83.0 &85.9 & 85.3 &77.2 \\

\rotatebox{0}{\textcolor{ncar}{$\blacksquare$} car} &\cellcolor[gray]{0.93}\underline{89.2} & 84.4 & 82.8 & 88.4 &\textbf{90.9} &80.9 \\

\rotatebox{0}{\textcolor{nconstruct}{$\blacksquare$} const. veh.}  & \cellcolor[gray]{0.93}\textbf{67.3} &\underline{47.3} & 38.3  &42.2 &35.1 &30.2 \\

\rotatebox{0}{\textcolor{nmotor}{$\blacksquare$} motorcycle}  &\cellcolor[gray]{0.93}{70.6} &{59.1}  &27.4  &\underline{72.4} &\textbf{77.5} &66.8  \\

\rotatebox{0}{\textcolor{npedestrian}{$\blacksquare$} pedestrian} & \cellcolor[gray]{0.93}{64.9} & 61.9 &{44.9}  & \textbf{72.2} &\underline{71.3} & 69.6  \\

\rotatebox{0}{\textcolor{ntraffic}{$\blacksquare$} traffic cone} &\cellcolor[gray]{0.93}{43.4} &{42.1} &24.0  &\textbf{63.1} & \underline{58.8} & 52.1 \\

\rotatebox{0}{\textcolor{ntrailer}{$\blacksquare$} trailer} &\cellcolor[gray]{0.93}\textbf{72.4} &58.8 & 55.4 & \underline{61.3} & 57.4 & 54.2 \\

\rotatebox{0}{\textcolor{ntruck}{$\blacksquare$} truck} &\cellcolor[gray]{0.93}\textbf{86.5} &\underline{82.8}& 73.6 &76.5 &76.1 & 72.3  \\

\rotatebox{0}{\textcolor{ndriveable}{$\blacksquare$} drive. suf.} &\cellcolor[gray]{0.93}{93.2} &{93.0} & 91.7  &\underline{96.0} &\textbf{96.5} & 94.1  \\

\rotatebox{0}{\textcolor{nother}{$\blacksquare$} other flat} &\cellcolor[gray]{0.93}\textbf{73.1} &{67.5} & 60.7 & 71.6 & \underline{71.1} & 66.6 \\

\rotatebox{0}{\textcolor{nsidewalk}{$\blacksquare$} sidewalk} &\cellcolor[gray]{0.93}{74.2} &{67.4} & 59.8 &\textbf{76.4}  & \underline{74.7} &63.5\\

\rotatebox{0}{\textcolor{nterrain}{$\blacksquare$} terrain} &\cellcolor[gray]{0.93}\underline{74.6} &{68.5} & 61.1 & \textbf{75.4}  & 74.0 & 70.1 \\

\rotatebox{0}{\textcolor{nmanmade}{$\blacksquare$} manmade} & \cellcolor[gray]{0.93}82.6 &81.0 & 78.2 & \underline{86.7} & \textbf{87.3} & 83.1 \\

\rotatebox{0}{\textcolor{nvegetation}{$\blacksquare$} vegetation} &\cellcolor[gray]{0.93}79.8 & 78.5 & 76.5 & \underline{84.4} & \textbf{85.7} &79.8  \\
 \bottomrule
\end{tabular}
}
\vspace{-0pt}
\label{tab_nus}
\end{center}
\end{table*}

 \begin{figure*}[!ht]
   \vspace{-0pt}
	\begin{center}
		\includegraphics[width=0.75\linewidth]{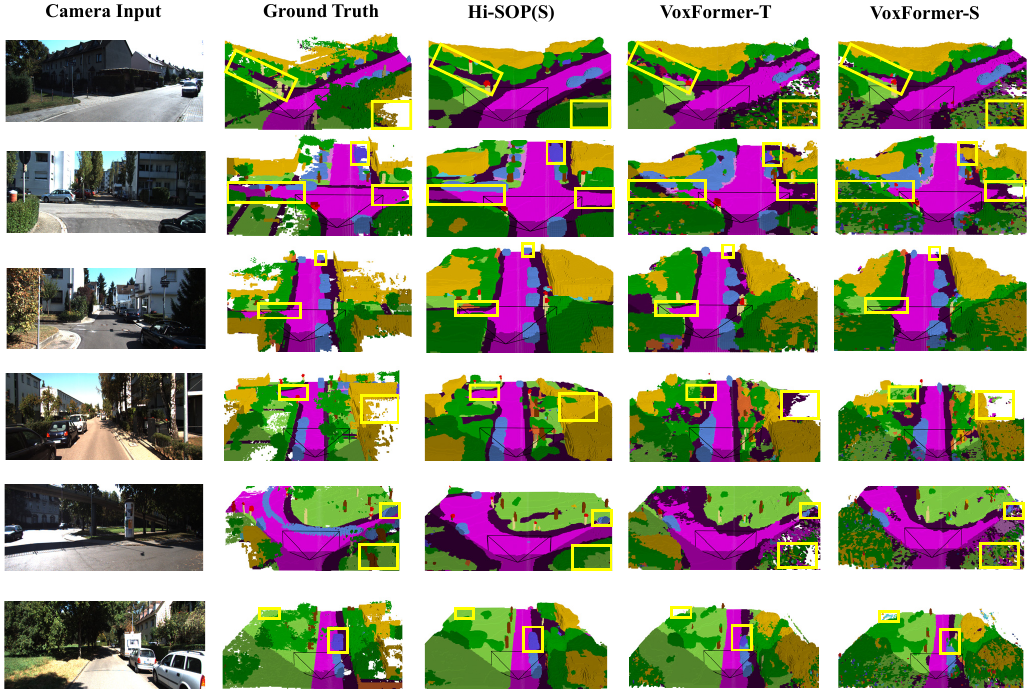}   
        \vspace{0pt}
		\begin{tabular}{cccccc}
			\multicolumn{6}{c}{
				\small
				\textcolor{bicycle}{$\blacksquare$}bicycle~
				\textcolor{car}{$\blacksquare$}car~
				\textcolor{motorcycle}{$\blacksquare$}motorcycle~
				\textcolor{truck}{$\blacksquare$}truck~
				\textcolor{other-vehicle}{$\blacksquare$}other-veh.~
				\textcolor{person}{$\blacksquare$}person~
				\textcolor{bicyclist}{$\blacksquare$}bicyclist~
				\textcolor{motorcyclist}{$\blacksquare$}motorcyclist~
				\textcolor{road}{$\blacksquare$}road~
				}
    \\
			\multicolumn{6}{c}{
				\small
                    \textcolor{parking}{$\blacksquare$}parking \textcolor{sidewalk}{$\blacksquare$}sidewalk \textcolor{other-ground}{$\blacksquare$}other.grd \textcolor{building}{$\blacksquare$}building \textcolor{fence} {$\blacksquare$}fence \textcolor{vegetation}{$\blacksquare$}vegetation \textcolor{trunk}{$\blacksquare$}trunk \textcolor{terrain}{$\blacksquare$}terrain \textcolor{pole}{$\blacksquare$}pole \textcolor{traffic-sign}{$\blacksquare$}traf.sign	
			}
		\end{tabular}	
 	\end{center}
        \vspace{-0pt}
	\caption{ \textbf{Qualitative results} of our method and others on the SemanticKITTI validation set. Our proposed Hi-SOP captures more complete and accurate scenery layouts compared with VoxFormer. Meanwhile, Hi-SOP hallucinates more proper scenery beyond the camera's field of view.\looseness=-1} 
	\label{fig_q}
 \vspace{-0pt}
\end{figure*}

\begin{figure*}[!ht]
   \vspace{-0pt}
	\begin{center}
		\includegraphics[width=0.75\linewidth]{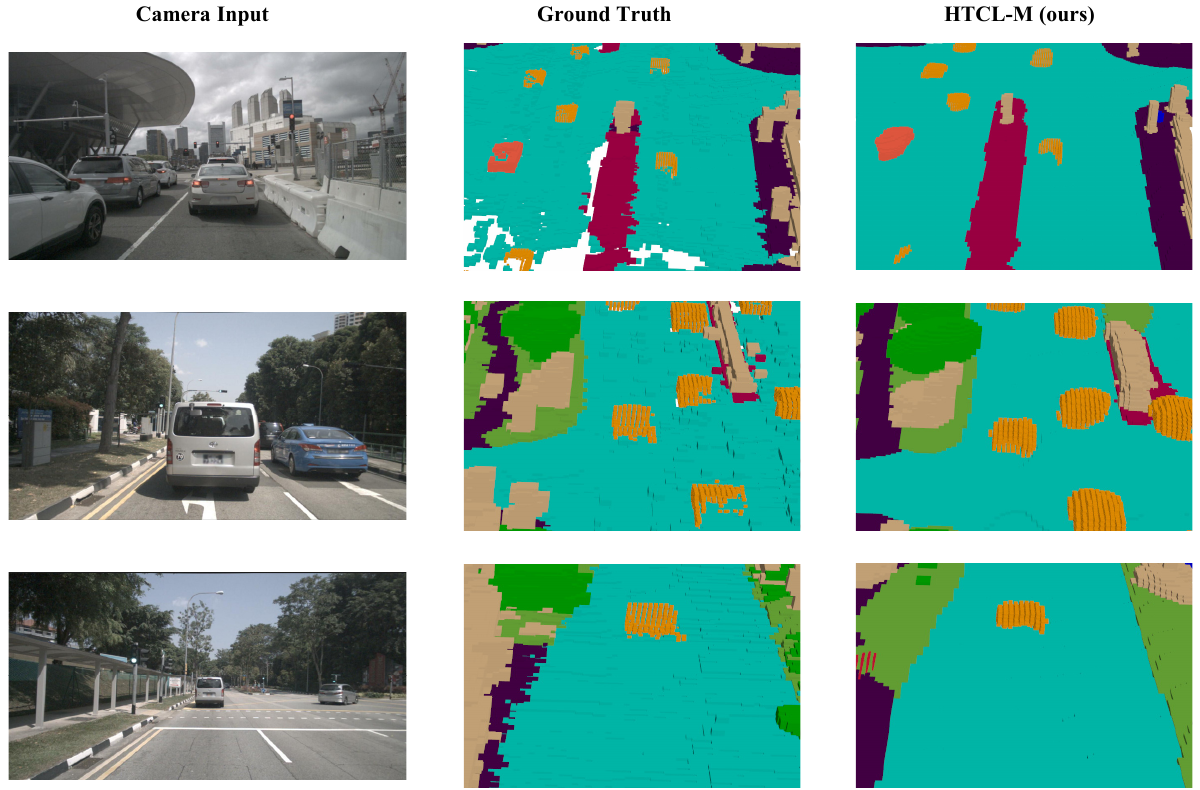}   
        \vspace{0pt}
		\begin{tabular}{cccccccc}
			\multicolumn{8}{c}{
				\small
				\textcolor{barrier1}{$\blacksquare$}barrier \textcolor{bicycle1}{$\blacksquare$}bicycle \textcolor{bus1}{$\blacksquare$}bus \textcolor{car1}{$\blacksquare$}car \textcolor{const. veh.1}{$\blacksquare$}const.veh. \textcolor{motorcycle1}{$\blacksquare$}motorcycle \textcolor{pedestrian1}{$\blacksquare$}pedestrian \textcolor{truck1}{$\blacksquare$}truck \textcolor{sidewalk1}{$\blacksquare$}sidewalk~
        
				}
    \\
			\multicolumn{7}{c}{
				\small

                \textcolor{traffic cone1}{$\blacksquare$}traffic cone \textcolor{trailer1}{$\blacksquare$}trailer \textcolor{drive. suf.1}{$\blacksquare$}drive.suf. \textcolor{other flat1}{$\blacksquare$}other flat \textcolor{terrain1}{$\blacksquare$}terrain \textcolor{manmade1}{$\blacksquare$}manmade \textcolor{vegetation1}{$\blacksquare$}vegetation~
			}
		\end{tabular}	
 	\end{center}
        \vspace{-5pt}
	\caption{ \textbf{Qualitative results} of our method and others on the NuScenes-Occupancy validation set. 
 Our proposed Hi-SOP can generate more complete and comprehensive semantic scenes compared with the ground truth.
 } 
	\label{fig_open}
 \vspace{-0pt}
\end{figure*}

\begin{figure*}[htbp]
   \vspace{-0pt}
	\begin{center}
		\includegraphics[width=0.75\linewidth]{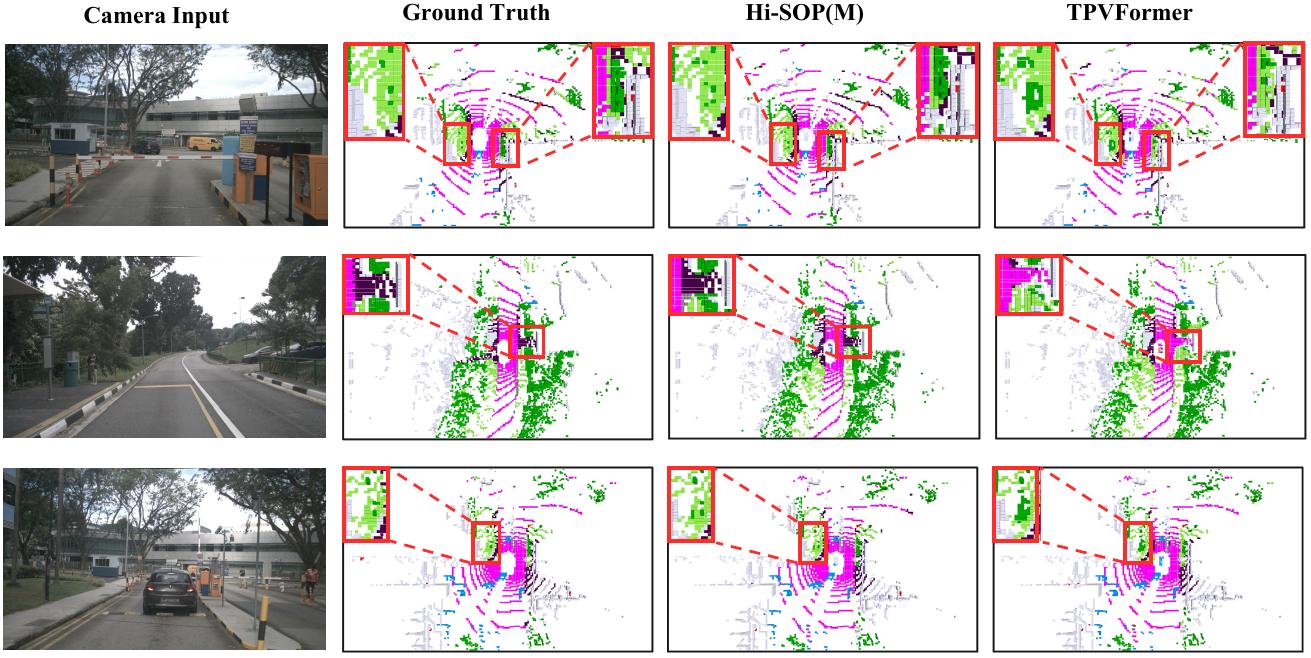}   
        \vspace{0pt}
				\begin{tabular}{cccccccc}
			\multicolumn{8}{c}{
				\small
				\textcolor{nbarrier}{$\blacksquare$}barrier \textcolor{nbicycle}{$\blacksquare$}bicycle \textcolor{nbus}{$\blacksquare$}bus \textcolor{ncar}{$\blacksquare$}car \textcolor{nconstruct}{$\blacksquare$}const.veh. \textcolor{nmotor}{$\blacksquare$}motorcycle \textcolor{npedestrian}{$\blacksquare$}pedestrian \textcolor{ntruck}{$\blacksquare$}truck \textcolor{nsidewalk}{$\blacksquare$}sidewalk~
        
				}
    \\
			\multicolumn{7}{c}{
				\small

                \textcolor{ntraffic}{$\blacksquare$}traffic cone \textcolor{ntrailer}{$\blacksquare$}trailer \textcolor{ndriveable}{$\blacksquare$}drive.suf. \textcolor{nother}{$\blacksquare$}other flat \textcolor{nterrain}{$\blacksquare$}terrain \textcolor{nmanmade}{$\blacksquare$}manmade \textcolor{nvegetation}{$\blacksquare$}vegetation~
			}
		\end{tabular}	
 	\end{center}
        \vspace{-0pt}
	\caption{ \textbf{Qualitative results} of our method and others on the NuScenes validation set. Our proposed Hi-SOP generates more accurate semantic labels compared with the results from TPVFormer.
 } 
	\label{fig_nus}
 \vspace{-0pt}
\end{figure*}

\section{Experiment}    

\subsection{Datasets and Metrics}
\noindent\textbf{SemanticKITTI.}
The SemanticKITTI dataset \cite{behley2019semantickitti} includes 22 outdoor scenes characterized by LiDAR scans and stereo images. The ground truth is structured into 256$\times$256$\times$32 voxel grids, with each voxel measuring 0.2m in all dimensions and annotated with 21 semantic classes (19 semantic, 1 free, and 1 unknown). Consistent with prior studies \cite{cao2022monoscene,li2023voxformer}, we divided the dataset into 10 training scenes, 1 validation scene, and 1 test scene. We evaluated our method using both stereo (HTCL-S) and monocular images Hi-SOP(M) on the SemanticKITTI.\looseness=-1

\noindent\textbf{NuScenes.}
The NuScenes dataset \cite{caesar2020nuscenes} is an autonomous driving dataset collected in Boston and Singapore. It comprises 1,000 driving sequences across various environments, with each sequence lasting approximately 20 seconds. Keyframes are annotated at a rate of 2Hz with 3D bounding boxes. The Panoptic NuScenes dataset~\cite{fong2022panoptic} extends the original NuScenes dataset by providing annotations for LiDAR semantic segmentation. Following previous works~\cite{huang2023tri,zhang2023occformer}, we utilize sparse LiDAR point supervision for 3D semantic occupancy prediction. We divided the dataset into training, validation, and testing splits containing 700, 150, and 150 scenes, respectively. 
Note that our monocular-based approach of Hi-SOP(M) is exclusively applied on the NuScenes dataset due to the absence of stereo images.\looseness=-1

\noindent\textbf{NuScenes-Occupancy.}
The NuScenes-Occupancy dataset \cite{wang2023openoccupancy} is an extension of the NuScenes dataset \cite{caesar2020nuscenes}, which provides dense semantic occupancy annotations for 850 scenes comprising 34,000 keyframes with 360-degree LiDAR scans. We divide the dataset into 28,130 training frames and 6,019 validation frames as described in \cite{wang2023openoccupancy}. Each frame includes 400K occupied voxels labeled with 17 semantic classes. 
We exclusively apply our monocular-based Hi-SOP(M) on the OpenOccupancy dataset as on the NuScenes dataset.\looseness=-1

\noindent\textbf{Evaluation Metrics.}
Following previous works~\cite{cao2022monoscene,li2023voxformer}, we adopt the mean Intersection over Union (mIoU) as the primary metric for evaluating the Semantic Scene Completion (SSC) task and the LiDAR semantic segmentation task. Additionally, the Intersection over Union (IoU) metric is used to evaluate the performance of the class-agnostic scene completion (SC) task.
For the evaluation of LiDAR semantic segmentation results, the LiDAR points are only used to query their corresponding semantic logits from the predicted 3D semantic occupancy volume following~\cite{huang2023tri,zhang2023occformer}.\looseness=-1

\subsection{Experimental Setup}\label{version_fast}
Following standard practices \cite{cao2022monoscene,zhang2023occformer,li2023voxformer}, we initialize the UNet backbone with pre-trained weights from EfficientNetB7~\cite{tan2019efficientnet}. By default, the model takes the current and previous three image frames as inputs. We implement our model on PyTorch with a batch size of 4 and train the model for 24 epochs using the AdamW optimizer \cite{loshchilov2017decoupled}. The learning rate is set at $1\times10^{-4}$, with a weight decay of 0.01.\looseness=-1

\subsection{Main Results
}
\subsubsection{Quantitative Comparison} We compare the quantitative results with the state-of-the-art camera-based semantic scene completion methods on the SemanticKITTI, NuScenes-Occupancy and NuScenes datasets, respectively.

As detailed in Table~\ref{tabq1} and Table~\ref{tabq1_}, we conducted a comparison analysis of our proposed method against existing best methods on the SemanticKITTI dataset, including VoxFormer~\cite{li2023voxformer}, OccFormer~\cite{zhang2023occformer}, SurroundOcc~\cite{wei2023surroundocc}, TPVFormer~\cite{huang2023tri}, and MonoScene~\cite{cao2022monoscene}. VoxFormer-T is a temporal baseline utilizing the current and previous four frames as inputs. Our method demonstrates superior performance, surpassing VoxFormer-T by 4.84 mIoU on the SemanticKITTI validation set and 4.08 mIoU on the SemanticKITTI test set, with fewer historical inputs (3 vs. 4).

Further quantitative evaluations on the NuScenes-Occupancy validation set are presented in Table~\ref{tab_open}. For depth map generation required by AICNet~\cite{li2021anisotropic} and 3DSketch~\cite{3d-sketch}, LiDAR points are projected and densified following~\cite{wang2023openoccupancy}. 
Despite the inherent advantage of LiDAR in IoU measurements due to its accurate 3D geometrical data, our method outperforms all competing methods in terms of mIoU, including those based on LiDAR like LMSCNet \cite{roldao2020lmscnet} and JS3C-Net \cite{yan2021sparse}. This demonstrates the robustness and effectiveness of Hi-SOP in the semantic occupancy prediction task.\looseness=-1

The quantitative results on the NuScenes validation set are presented in Table~\ref{tab_nus}. We compare our method with state-of-the-art camera-based methods of 
OccFormer~\cite{zhang2023occformer} and TPVFormer~\cite{huang2023tri}, and LiDAR-based methods of SalsaNext~\cite{cortinhal2020salsanext}, PolarNet~\cite{zhang2020polarnet} and RangeNet++~\cite{milioto2019rangenet++}. 
Despite LiDAR's inherent advantage in accurate 3D geometric measurements, 
our method outperforms all the other methods in terms of mIoU.\looseness=-1

\begin{table}[!ht]
\vspace{-0pt}
\begin{center}
\renewcommand\tabcolsep{12.0pt}
\scriptsize
\caption{ \textbf{Evaluation results} of temporal stereo variants on the SemanticKITTI validation set.  
The {``S-T''} and {``M-T''} denote temporal stereo images and temporal monocular images, respectively. 
For MonoScene${}^{\ddagger}$, TPVFormer${}^{\ddagger}$ and OccFormer${}^{\ddagger}$, we employ stacked temporal stereo images as inputs following VoxFormer-T.}
\resizebox{0.99\linewidth}{!}{
\begin{tabular}{l|ccc} 
\toprule
 \textbf{Methods}& \textbf{Input}   & \textbf{mIoU(\%)} $\uparrow$ & \textbf{Time(s)} $\downarrow$      \\  \midrule

MonoScene${}^{\ddagger}$ &  S-T  & 12.96 & \textbf{0.281}  \\
TPVFormer${}^{\ddagger}$    &  S-T   &   13.21 & 0.324   \\
OccFormer${}^{\ddagger}$   & S-T  &  13.57 & 0.348   \\

VoxFormer-T   &  S-T  & 13.35 & 0.307    \\ \midrule
\rowcolor{gray!10} Hi-SOP(M)  & M-T & 16.63  & 0.294  \\ 
\rowcolor{gray!10} Hi-SOP(S)  & S-T &\textbf{18.19} & {0.302}   \\ 
 \bottomrule
\end{tabular}
}
\vspace{-0pt}
\label{tabq3}
\end{center}
\end{table}

\subsubsection{Qualitative Comparison} We present comparative analyses of our qualitative results against other state-of-the-art camera-based semantic scene completion methods on the SemanticKITTI, NuScenes-Occupancy, and NuScenes datasets, respectively.\looseness=-1

Figure~\ref{fig_q} presents the qualitative comparison between our proposed method and VoxFormer~\cite{li2023voxformer} on the SemanticKITTI validation set. As we can see from the figure, the real-world scenes are inherently complex, and the sparsity of the annotated ground truth presents significant challenges in fully reconstructing semantic scenes from limited visual cues. 
Our method surpasses VoxFormer in capturing a more complete and accurate layout of the scenery, as illustrated by the crossroads in the first and third rows. 
Additionally, our method effectively infers the scenery beyond the camera's field of view, notably in shadowed areas shown in the first and fifth rows, and exhibits marked improvements in handling dynamic objects, such as trucks in the second and sixth rows.

Furthermore, Figure~\ref{fig_open} illustrates the prediction results of our method on the NuScenes-Occupancy validation set.
Our proposed method generates much denser and more realistic results compared
with the ground truth.\looseness=-1

The qualitative results on the NuScenes validation set are presented in Figure~\ref{fig_nus}.
Following previous works~\cite{huang2023tri}, we use only RGB
images as input, while the LiDAR points are only used to query
their features and for supervision in the training phase.
Our proposed method generates more accurate semantic labels compared with the results from TPVFormer~\cite{huang2023tri}. \looseness=-1

\subsubsection{Temporal Stereo Variants Evaluation.}
To ensure a fair and comprehensive comparison, we have implemented temporal stereo variants of baseline models as detailed in Table~\ref{tabq3}. Following the approach of VoxFormer-T, we utilize stacked temporal stereo images as inputs, creating variants of MonoScene${}^{\ddagger}$~\cite{cao2022monoscene}, TPVFormer${}^{\ddagger}$~\cite{huang2023tri}, and OccFormer${}^{\ddagger}$~\cite{zhang2023occformer}. It is important to note that whereas VoxFormer-T~\cite{li2023voxformer} originally utilizes four previous frames, the stereo variants we developed employ only three previous frames, aligning with our method. As demonstrated in the table, our approach consistently achieves superior performance using the same temporal inputs.

\begin{table*}[htbp]\centering
\vspace{-0pt}
\renewcommand\tabcolsep{10.2pt}
\scriptsize
\caption{\textbf{Ablation study} for different architectural components on the SemanticKITTI validation set. The full names of different components are in Sec IV-D.\looseness=-1}
\vspace{-0pt}
\resizebox{0.99\textwidth}{!}{
\begin{tabular}{c|cc|cc|cc|c|cc} \toprule
\multirow{2}{*}{\textbf{GCL}} & \multicolumn{2}{c|}{\textbf{TVC}}                     & \multicolumn{2}{c|}{\textbf{CPA}}                     & \multicolumn{2}{c|}{\textbf{ADR}}                     & \multirow{2}{*}{\textbf{DHBT}} & 
\multirow{2}{*}{\textbf{IoU(\%) $\uparrow$}} & \multirow{2}{*}{\textbf{mIoU(\%) $\uparrow$}} \\
& \begin{tabular}[c]{@{}c@{}}Feature\\Volume\end{tabular}
  & \begin{tabular}[c]{@{}c@{}}Cost\\Volume\end{tabular}   & \begin{tabular}[c]{@{}c@{}}Scale-aware \\Isolation\end{tabular}   & \begin{tabular}[c]{@{}c@{}}Multi-\\group\end{tabular}   & Affinity & Deformable    &   &   \\ \midrule
  &\checkmark &  &\checkmark &\checkmark &\checkmark & \checkmark &\checkmark  & 43.17 & 16.94 \\ \midrule
  
\checkmark& &\checkmark  &\checkmark &\checkmark &\checkmark & \checkmark &\checkmark  &  44.05  &17.10  \\ \midrule

\checkmark&\checkmark &  & &\checkmark & \checkmark &\checkmark &\checkmark  &43.16  &  16.25 \\
\checkmark&\checkmark &  & \checkmark& & \checkmark &\checkmark &\checkmark   & 43.22 &  16.33\\  \midrule

\checkmark&\checkmark &  &\checkmark &\checkmark &  & \checkmark&\checkmark  & 42.85 & 15.74\\
\checkmark&\checkmark &  &\checkmark &\checkmark & \checkmark & &\checkmark   & 43.03 & 16.21 \\  \midrule

\checkmark&\checkmark &  &\checkmark &\checkmark &\checkmark  & \checkmark  & & 44.16 & 17.05 \\  \midrule
\rowcolor{gray!10}  \checkmark&\checkmark & &\checkmark &\checkmark &\checkmark & \checkmark & \checkmark & \textbf{45.56} & \textbf{18.19}  \\ \bottomrule
\end{tabular}
}
\vspace{-0pt}
\label{tab_ar}
\end{table*}

\begin{table}[htbp]\centering
\vspace{-0pt}
\renewcommand\tabcolsep{1.0pt}
\scriptsize
\caption{\textbf{Ablation studies} of quantity setting for the Multi-group Context Generation module and the Multi-level Deformable Block.}
\resizebox{0.98\linewidth}{!}{
\begin{tabular}{ccc|ccc|cc} \toprule
\multicolumn{3}{c|}{\textbf{Context Generation}} & \multicolumn{3}{c|}{\textbf{Deformable Block}} & \multirow{2}{*}{\textbf{mIoU(\%) $\uparrow$}} & \multirow{2}{*}{\textbf{Time(s) $\downarrow$}} \\

\qquad 1   &\qquad 3  &\qquad 5 \qquad & \qquad 1    & \qquad 3    &\qquad 5  \qquad & & \\  \midrule 
\qquad \checkmark & & & & \qquad \checkmark & & 16.33 &  0.289     \\ 
\qquad &\qquad &\qquad  \checkmark  & & \qquad \checkmark &  & 18.26 &  0.318 \\   

 \midrule

& \qquad \checkmark &  & \qquad \checkmark & \qquad & & 17.59 & 0.291 \\ 

& \qquad \checkmark &  & \qquad  & \qquad  & \qquad \checkmark & 18.23  & 0.316  \\  \midrule
\rowcolor{gray!10} \qquad  & \qquad \checkmark &\qquad  & \qquad  &\qquad \checkmark & \qquad  &{18.19} &  {0.302} \\  \bottomrule
    \end{tabular}
}
\vspace{-0pt}
\label{tab_ab}
\end{table}

\subsection{Ablation Study}\label{sec_ab}
We conduct comprehensive ablation tests for our proposed method using the SemanticKITTI validation set. Specifically, we evaluate the effects of different architectural components in Table~\ref{tab_ar} and analyze the role of temporal inputs in Table~\ref{tab_tem}.
Moreover, we conduct ablation studies of quantity setting for the Multi-group Context Generation and the Multi-level Deformable Block, as presented in Table~\ref{tab_ab}.\looseness=-1

\noindent \textbf{Effect of GCL.}
The ablation results for Geometric Confidence-aware Lifting (GCL) are presented in the second row of Table~\ref{tab_ar}. 
As shown in the table, replacing the typical lifting process as in previous work~\cite{philion2020lift,zhang2023occformer} with our proposed Geometric Confidence-aware Lifting increases the IoU and mIoU by 2.39 and 1.25, respectively. We attribute such improvements to the explicit geometric modeling with depth distribution confidence.

\noindent \textbf{Effect of TVC.}
The ablation results for Temporal Volume Construction (TVC) are presented in the third row of Table~\ref{tab_ar}. 
Replacing the cost volume with the feature volume notably enhances performance, increasing the IoU and mIoU by 1.51 and 1.09, respectively. The enhancement is attributed to the preservation of fine-grained feature context.\looseness=-1

\noindent \textbf{Effect of CPA.}
Details on the ablation of Cross-frame Pattern Affinity (CPA) are shown in the fourth and fifth rows of Table~\ref{tab_ar}. Enhancing the original cosine similarity with scale-aware isolation and incorporating multi-group context generation significantly improves mIoU, with increases of 1.94 and 1.86, respectively.

\noindent \textbf{Effect of ADR.}
The ablation study for Affinity-based Dynamic Refinement (ADR) involved removing the affinity weights and replacing deformable convolutions with standard convolutions, as shown in the sixth and seventh rows of Table~\ref{tab_ar}. Utilizing affinity information proved effective in modeling contextual correspondences, resulting in notable performance improvements of 2.71 IoU and 2.45 mIoU. Additionally, dynamic refinement through deformable convolutions facilitates efficient and flexible contextual modeling, further enhancing IoU and mIoU by 2.53 and 1.98, respectively.

\noindent \textbf{Effect of DHBT.}
The ablation study on Depth-Hypothesis-Based Transformation (DHBT) is depicted in the eighth row of Table~\ref{tab_ar}. For comparative analysis, we remove the module and directly fuse the temporal volume and the voxel feature volume with naive concatenation. 
As we can see, the depth-hypothesis-based transformation yields substantial improvements in IoU and mIoU, with increases of 1.40 and 1.14, respectively.\looseness=-1

\noindent \textbf{Module Quantity Setting.}
We conduct ablation studies of quantity setting for the Multi-group Context Generation and the Multi-level Deformable Block, as presented in the ninth row of Table~\ref{tab_ab}.
As introduced in Section~\ref{cpa}, we employ multiple groups of contextual features to facilitate diverse independent similarity learning.
The results in Table~\ref{tab_ab} demonstrate that leveraging 3 contextual groups yields a significant performance improvement, while employing more groups (5 groups) leads to a relatively slight improvement.
Similarly, the enhancement of utilizing more feature levels (5 levels) in the Multi-level Deformable Block is also relatively minor.
Therefore, considering the time consumption and parameter efficiency, we adopt 3 contextual groups in the Multi-group Context Generation and 3 feature levels in the Multi-level Deformable Block as the default settings.\looseness=-1

\begin{table}[!t]
\vspace{-0pt}
\begin{center}
\scriptsize
\caption{{Effect of using a different number of temporal frames.} These models are evaluated on the SemanticKITTI validation set.}
\renewcommand\tabcolsep{4.1pt}
\scriptsize
\resizebox{1.0\linewidth}{!}{
\begin{tabular}{ccccc|cc}\toprule
\multicolumn{5}{c|}{\textbf{Temporal Inputs}} &\multirow{2}{*}{\textbf{mIoU(\%)} $\uparrow$} &\multirow{2}{*}{\textbf{Time(s)} $\downarrow$}   \\
 $I^{rgb}_{t-1}$ & $I^{rgb}_{t-2}$ & $I^{rgb}_{t-3}$ &$I^{rgb}_{t-4}$ &$I^{rgb}_{t-5}$     \\ \midrule
\checkmark & &  &  & & 15.14 & 0.273   \\
 \checkmark&\checkmark &  & & & 16.58 & 0.287 \\
\rowcolor{gray!10}  \checkmark  & \checkmark &\checkmark &&  &  18.19  & 0.302   \\
 \checkmark  & \checkmark &\checkmark &\checkmark&     & 18.36  & 0.315 \\
 \checkmark  & \checkmark &\checkmark &\checkmark &\checkmark  & 18.45 & 0.328 \\ \bottomrule
\end{tabular}
}
\vspace{-0pt}
\label{tab_tem}
\end{center}
\end{table}

\noindent \textbf{Temporal Inputs.}
The effect of using a different number of temporal frames in terms of prediction performance and inference speed are outlined in Table~\ref{tab_tem}. The results in indicate that the marginal gains in effectiveness when using more than three previous frames are minimal compared to the increase in computational time. Therefore, we have chosen three frames as our standard configuration to achieve an optimal balance between efficiency and effectiveness.\looseness=-1

\section{Conclusion}
In this paper, we introduce Hi-SOP, a hierarchical context learning paradigm for semantic occupancy prediction with the disentanglement-before-composition scheme. 
For geometric context learning, to explicitly model the geometric information with the corresponding depth distribution confidence, we propose a geometric confidence-aware lifting module for reliable volumetric feature establishment.
For temporal context learning, Hi-SOP incorporates pattern affinity to model the contextual correspondence between current and historical frames. 
Subsequently, to dynamically compensate for incomplete observations, we propose to adaptively refine the feature sampling locations based on the initially high-affinity locations and their neighboring relevant regions.
Finally, the temporal context and the geometric context are aligned into a unified space, which are finally aggregated for reliable composition.
Our framework demonstrates superior performance over existing state-of-the-art camera-based methods and surpasses LiDAR-based methods in the semantic scene completion and LiDAR semantic segmentation tasks. 
We hope Hi-SOP could inspire further exploration in camera-based semantic occupancy prediction and enhance applications in 3D visual perception.\looseness=-1

\section*{Acknowledgments}
This work was supported in part by NSFC 62302246 and ZJNSFC under Grant LQ23F010008, and supported by High Performance Computing Center at Eastern Institute of Technology, Ningbo, and Ningbo Institute of Digital Twin.\looseness=-1

\bibliographystyle{IEEEtran}
 
\bibliography{egbib}

\begin{IEEEbiography}[{\includegraphics[width=1in,height=1.25in,clip,keepaspectratio]{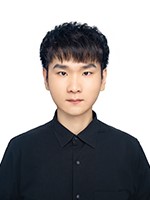}}]{Bohan Li} received the B.E. degree from the School of Control Engineering, Northeastern University (NEU),
Shenyang, China, in 2019. He received the M.E. degree from the School of Control Science and Engineering, South China University of Technology
(SCUT), Guangzhou, China, in 2022.

He is currently pursuing the Ph.D. degree in Shanghai Jiao Tong University (SJTU) and Eastern Institute of Technology (EIT). His research interests include 3D visual perception, robotics, and multi-modality content generation.
\end{IEEEbiography}
\vspace{-3em}

\begin{IEEEbiography}[{\includegraphics[width=1in,height=1.40in,clip,keepaspectratio]{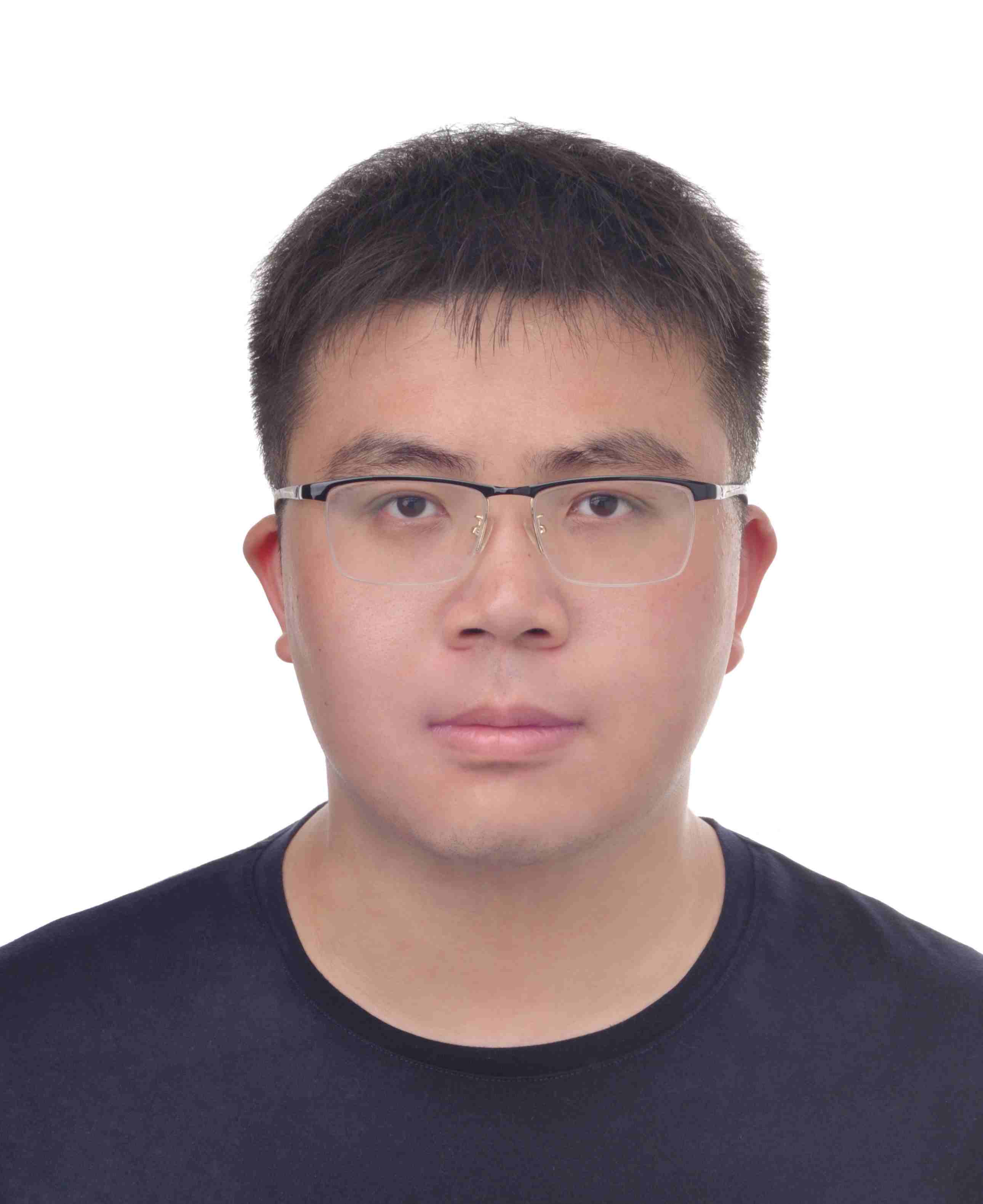}}]{Jiajun Deng} is a research fellow at the University of Adelaide, Australian Institute for Machine Learning. He received his Ph.D. degree (2021) and a B.E degree (2016) from the department of Electrical Engineering and Information Science at the University of Science and Technology of China.  He served as a guest editor of the Special Issue “Pre-trained Models for Multi-modality Understanding” of IEEE Transactions on Multimedia, in 2023. He also served as the Area Chair for ACM Multimedia, in 2024. His research interests include computer vision, multi-modality understanding, and embodied AI.
\end{IEEEbiography}
\vspace{-3em}

\begin{IEEEbiography}[{\includegraphics[width=1in,height=1.25in,clip,keepaspectratio]{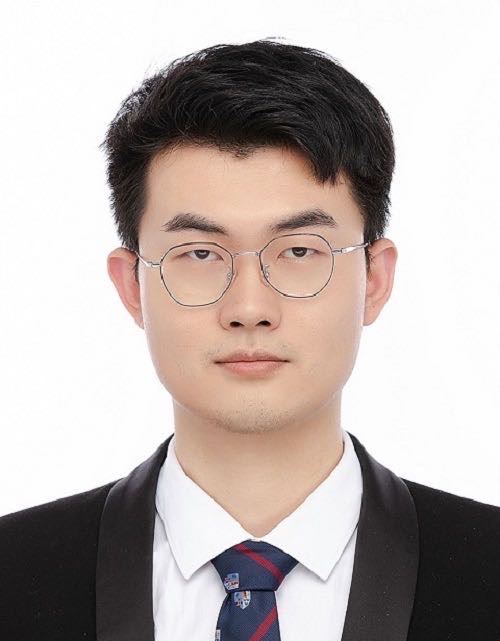}}]{Yasheng Sun} received the B.E. degree from Nanjing University of Aeronautics and Astronautics, Nanjing, China, in 2017. He received the M.E. degree from the School of Mechanical Engineering, Shanghai Jiao Tong University, Shanghai, China, in 2020. 
He received the Ph.D. degree in Computer Science from the School of Computing, Tokyo Institute of Technology, Japan, in 2024.

His current research interest includes cross-modal generation, 3D generative model, stable diffusion model and its application in computer vision.
\end{IEEEbiography}
\vspace{-3em}

\begin{IEEEbiography}[{\includegraphics[width=1in,height=1.25in,clip,keepaspectratio]{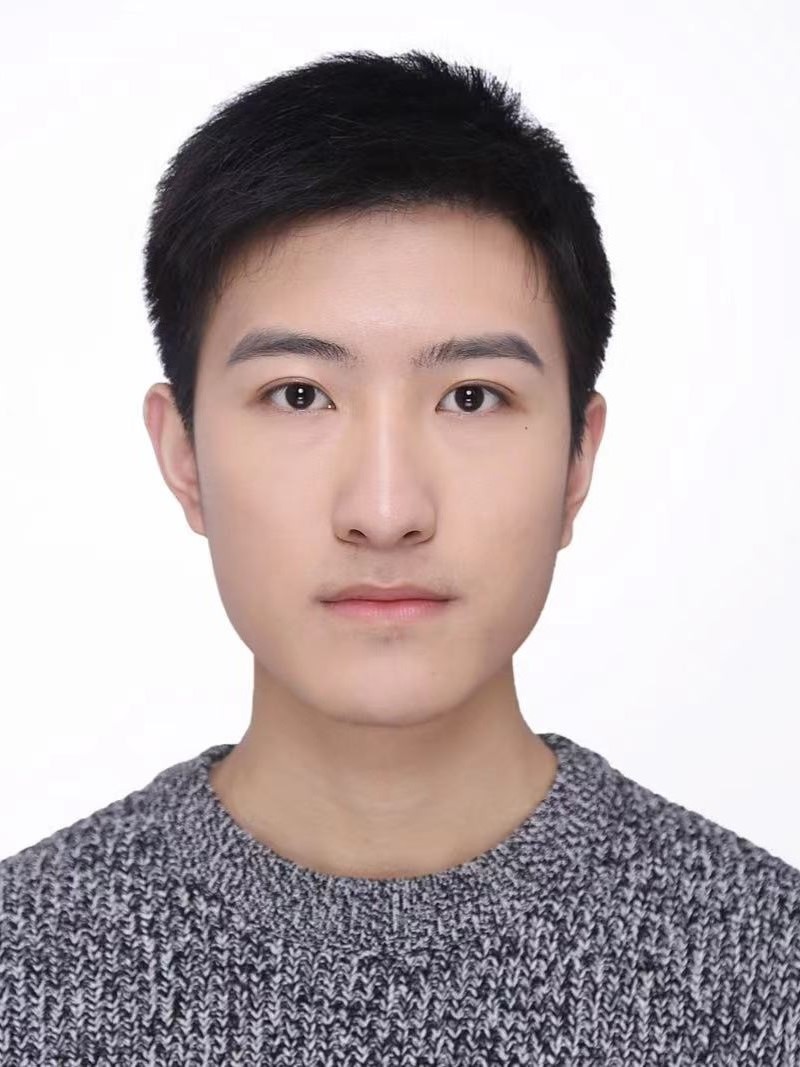}}]{Xiaofeng Wang} received the B.E. degree from the School of Automation, Nanjing University of Science and Technology (NJUST), Nanjing, China, in 2020. He is currently pursuing the Ph.D. degree in Institute of Automation, Chinese Academy of Science (CASIA), Beijing, China. 

His current research areas include 3D perception and video generation. He has co-authored 10+ journal and conference papers mainly on computer vision autonomous-driving problems, including CVPR, ECCV, ICCV, AAAI, and ICLR.

\end{IEEEbiography}
\vspace{-3em}

\begin{IEEEbiography}[{\includegraphics[width=1in,height=1.25in,clip,keepaspectratio]{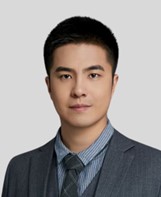}}]{Xin Jin} has been a tenure track Assistant Professor with the Eastern Institute of Technology (EIT), Ningbo, China. He is also a Researcher at the Ningbo Institute of Digital Twin. He received his Ph.D. degree in Electronic Engineering and Information Science from the University of Science and Technology of China (USTC). His research interests include computer vision, intelligent media computing, and deep learning. He has over 10 granted patent applications, around 40 publications, and over 3,500 Google citations. He is an IEEE member, and reviewer of IEEE Transactions on Image Processing (TIP), IEEE Transactions on Multimedia (TMM), and IEEE Transactions on Circuits and Systems for Video Technology (TCSVT).
\end{IEEEbiography}
\vspace{-3em}

\begin{IEEEbiography}[{\includegraphics[width=1in,height=1.25in,clip,keepaspectratio]{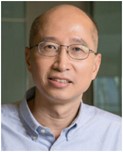}}]{Wenjun Zeng} (Fellow, IEEE) received the B.E. degree from Tsinghua University, Beijing, China,
in 1990, the M.S. degree from the University of Notre Dame, Notre Dame, IN, USA, in 1993, and the Ph.D. degree from Princeton University, Princeton, NJ, USA, in 1997. He has been a Chair Professor and the Vice President for Research at the Eastern Institute for Advanced Study (EIAS) / Eastern Institute of Technology (EIT), Ningbo, China, since October 2021. He is also the founding Executive Director of the Ningbo Institute of Digital Twin. He was a Sr. Principal Research Manager and a member of the Senior Leadership Team at Microsoft Research Asia, Beijing, from 2014 to 2021, where he led the video analytics research empowering the Microsoft Cognitive Services, Azure Media Analytics Services, Office, and Windows Machine Learning. He was with University of Missouri, Columbia, MO, USA from 2003 to 2016, most recently as a Full Professor. Prior to that, he had worked for PacketVideo Corp., Sharp Labs of America, Bell Labs, and Panasonic Technology. He has contributed significantly to the development of international standards (ISO MPEG, JPEG2000, and OMA).
Dr. Zeng is on the Editorial Board of the International Journal of Computer Vision. He was an Associate Editor-in-Chief of the IEEE Multimedia Magazine and an Associate Editor of the IEEE TRANSACTIONS ON CIRCUITS AND SYSTEMS FOR VIDEO TECHNOLOGY, IEEE TRANSACTIONS ON INFORMATION FORENSICS AND SECURITY, and IEEE TRANSACTIONS
ON MULTIMEDIA (TMM). He was on the Steering Committee of IEEE TRANSACTIONS ON MOBILE COMPUTING and IEEE TMM. He served as the Steering Committee Chair of IEEE ICME in 2010 and 2011, and has served as the General Chair or TPC Chair for several IEEE conferences (\textit{e.g.}, ICME’2018, ICIP’2017). He was the recipient of several best paper awards.\looseness=-1
\end{IEEEbiography}
\vspace{-3em}


\end{document}